\RequirePackage[2020-02-02]{latexrelease}
\documentclass[article]{clv3}
\usepackage[export]{adjustbox}
\usepackage{multirow}
\usepackage{multicol}
\usepackage{verbatim}
\usepackage[inline]{enumitem}
\usepackage{dirtytalk}
\usepackage{pgfplots}
\usepackage{tikz}
\usetikzlibrary{shapes,backgrounds}
\usetikzlibrary{shapes,arrows,calc}
\usepgfplotslibrary{statistics}
\usepackage{url}
\usepackage{float}
\usepackage{comment}
\usepackage{anyfontsize}
\usetikzlibrary{tikzmark}
\usetikzlibrary{calc}
\usepackage{caption}
\usepackage{makecell}
\usepackage{subcaption}
\usepackage{alphalph}
\usepackage{balance}

\usepackage{times}
\usepackage{latexsym}
\usepackage{subcaption,graphicx}
\usepackage{microtype}
\usepackage{tcolorbox}
\usepackage{url}
\usepackage{mathtools}
\usepackage{amsmath,amssymb}
\usepackage{cleveref}
\usepackage[perpage]{footmisc}

\usepackage{makecell}
\usepackage{lscape} 
\usepackage{microtype}
\usepackage{enumitem,kantlipsum}
\usepackage{adjustbox}
\usepackage{xcolor}
\usepackage{pifont}
\restylefloat{table}
\newcommand{\cmark}{\ding{51}}%
\newcommand{\xmark}{\ding{55}}%
\pgfplotsset{compat=1.18}

\crefformat{section}{\S#2#1#3} 
\crefformat{subsection}{\S#2#1#3}

\begin{document}
\issue{1}{1}{2023}

\runningtitle{On Bias and Fairness in NLP}

\runningauthor{Fatma Elsafoury}


\title{On Bias and Fairness in NLP: Investigating the Impact of Bias and Debiasing in Language Models on the Fairness of Toxicity Detection}

\author{Fatma Elsafoury}
\affil{Fraunhofer Research institute \\
Weizenbaum Research institute
}

\author{Stamos Katsigiannis}
\affil{Durham University}

\maketitle
\begin{abstract}
Language models are the new state-of-the-art natural language processing (NLP) models and they are being increasingly used in many NLP tasks. Even though there is evidence that language models are biased, the impact of that bias on the fairness of downstream NLP tasks is still understudied. Furthermore, despite that numerous debiasing methods have been proposed in the literature, the impact of bias removal methods on the fairness of NLP tasks is also understudied. In this work, we investigate three different sources of bias in NLP models, i.e. representation bias, selection bias and overamplification bias, and examine how they impact the fairness of the downstream task of toxicity detection. Moreover, we investigate the impact of removing these biases using different bias removal techniques on the fairness of toxicity detection. Results show strong evidence that downstream sources of bias, especially overamplification bias, are the most impactful types of bias on the fairness of the task of toxicity detection. We also found strong evidence that removing overamplification bias by fine-tuning the language models on a dataset with balanced contextual representations and ratios of positive examples between different identity groups can improve the fairness of the task of toxicity detection. Finally, we build on our findings and introduce a list of guidelines to ensure the fairness of the task of toxicity detection. 
\end{abstract}

\section{Introduction}
Language models (LM) are being used in different natural language processing (NLP) tasks, like in search engines \citep{Zhu2023LargeLM}, email filtering \citep{ABDULNABI2021853}, and content moderation \citep{elsafoury2021}.  Recent research has indicated that LMs are prone to learning and reproducing harmful social biases \cite{Garg2017, Caliskan2017, sweeney2019, nangia-etal-2020-crows, nadeem-etal-2021-stereoset}. For example, \citet{elsafoury_sos_2022} demonstrate that different static word embeddings learn to associate between marginalized identities (e.g., Women, LGBTQ, and non-white ethnicity) and profane words. Similar results have been found in contextual word embeddings as well \citep{ousidhoum-etal-2021-probing, nozza-etal-2022-measuring,Elsafoury2023SystematicOS}. 
However, the impact of social bias in LMs on downstream NLP tasks like toxicity detection is still understudied. For example, even though it has been shown in the literature that different bias metrics return different results \cite{badilla2020,elsafoury-etal-2022-comparative}, most of the studies that investigated the impact of bias in LMs on downstream NLP tasks used only one bias metric \cite{steed-etal-2022-upstream,goldfarb-tarrant-etal-2021-intrinsic}, which leaves the findings inconclusive. 

Understanding the impact of social bias on downstream tasks like toxicity detection is crucial, especially with research demonstrating that content written by marginalized identities is sometimes falsely flagged as toxic or hateful \cite{sap2019}. 
Furthermore, various methods have been proposed in the literature for removing bias from LMs and for ensuring that NLP tasks do not discriminate between different identity groups based on sensitive attributes like gender, ethnicity, religion, or sexual orientation \cite{zhao-etal-2017-men, liang2020towards, Webster-etal-2020-cda, meade-etal-2022-emprical-survey-debias, qian-etal-2022-perturbation}. Yet the impact of these debiasing methods on downstream NLP tasks is also understudied.

In this work, we aim to address these research gaps by investigating the impact of bias and debiasing in LMs on the downstream task of toxicity detection. \citet{lopez2021bias} distinguishes between three types of bias: Technological bias, Socio-Technical bias, and Societal bias based on the legal anti-discrimination regulations. In this study, we focus on socio-technical bias which is described as \textit{''A systematic deviation due to structural inequalities.''} 
In particular, we investigate three different sources of socio-technical bias from the predictive bias framework for NLP: (a) \textit{Representation bias}, (b) \textit{Selection bias}, and (c) \textit{Overamplification bias} \cite{shah-etal-2020-predictive, hovy2021five}. It must be noted that representation bias is also referred to as \textit{upstream} bias in the literature \cite{steed-etal-2022-upstream}, while selection and overamplification bias are refereed to as \textit{downstream} bias \cite{steed-etal-2022-upstream, badilla2020}. 

We use different metrics from the literature to measure representation bias and fairness to evaluate the consistency of our results. We also propose two metrics to measure selection and overamplification bias. First, we investigate the aforementioned sources of bias and their impact on the fairness of the downstream task of toxicity detection. Then, we remove these sources of bias and investigate whether their removal improves the fairness of toxicity detection. The aim of this work is to find the most impactful sources of bias and the most effective bias removal techniques to use to improve the fairness of toxicity detection without 
compromising its performance. To this end, this work aims to answer the following research questions:
\begin{enumerate}
\setlength\itemsep{-0.1em}
\item What is the impact of the different sources of bias on the fairness of the downstream task of toxicity detection? 
\item What is the impact of removing the different sources of bias on the fairness of the downstream task of toxicity detection? 
\item Which debiasing technique to use to ensure the fairness of the task of toxicity detection?
\item How to ensure the fairness of the toxicity detection task?
\end{enumerate}

To answer our research questions, we use statistical definitions of bias and fairness from the NLP literature to measure bias and fairness in three LMs: AlBERT-base-v2 \cite{albert}, BERT-base-uncased \cite{DBLP:conf/naacl/DevlinCLT19}, and RoBERTa-base \cite{roberta}.  We measure three sources of bias, i.e. representation bias (\cref{sec:intrinsic_bias}), selection bias (\cref{sec:selection_bias}), overamplification bias (\cref{sec:overap_bias}), using different metrics and investigate their impact on toxicity detection using statistical correlation. Then, we use different methods for representation bias removal (\cref{sec:sentDebias}), selection bias removal (\cref{sec:selection-bias-removal}) and overamplification bias removal (\cref{sec:over-bias-removal}), and investigate the impact of these bias removal methods on the models' fairness and performance. We then analyse our results to answer the first research question and to understand the impact of the different sources of bias on the models' fairness on toxicity detection (\cref{sec:RQ1}), as well as to understand the impact of removing different sources of bias on the fairness of toxicity detection and to answer our second research question (\cref{sec:RQ2}).
Thereafter, we further analyze the debiasing (bias removal) results to identify the most effective technique to ensure the models' fairness on the downstream task of toxicity detection and to answer our third research question (\cref{sec:counterfactual_fairness}). Finally, to answer the fourth research question, we built on the findings of this work on the fairness of toxicity detection and propose a list of guidelines to ensure the fairness of toxicity detection (\cref{sec:improving_fainress}). 
\\\\
\textbf{The main contributions of this paper can be summarized as follows:} 
\begin{enumerate}
    \item We provide a comprehensive investigation of different sources of bias in NLP models and their impact on the fairness of the task of toxicity detection.
    \item We provide a comprehensive investigation of different bias removal (debiasing) methods to remove different sources of bias and their impact on the fairness and performance of the task of toxicity detection.  
    \item We provide guidelines to ensure the fairness of the task of toxicity detection.
\end{enumerate}

Our findings suggest that the dataset used in measuring fairness impacts the measured fairness scores of toxicity detection, and using a balanced dataset improves the fairness scores. Unlike the findings of previous research \cite{goldfarb-tarrant-etal-2021-intrinsic}, our findings suggest that upstream (representation) bias is impactful on the fairness of the toxicity detection. However, our results suggest that downstream sources of bias (selection and overamplification) are more impactful on the fairness of the toxicity detection, which is in line with the findings of previous research \cite{steed-etal-2022-upstream}. However, unlike the findings of \cite{steed-etal-2022-upstream}, our results suggest that removing overamplification bias in the training dataset before fine-tuning is the most effective downstream bias removal method and improved the fairness of the toxicity detection. Our results show strong evidence that the most effective methods to remove overamplification bias is to use the simple method of lexical word replacement to create perturbations to balance the representation of the different identity groups in the training dataset. Finally, we build on these findings and provide a list of guidelines to check against to ensure the fairness of the task of toxicity detection.

Improving the fairness of toxicity detection is very critical for ensuring that the decisions made by the models are not based on sensitive attributes like race or gender. For transparency and to encourage further investigation, we make the code of this work publicly available\footnote{This link will be available upon acceptance.}. 

\section{Background}
In this section, we review the literature on the definition of \textit{Bias} and \textit{fairness} and the different metrics used in the literature to measure these two concepts. Then, we provide a critical review of the relevant work in the literature on investigating the impact of bias in LMs on downstream NLP tasks, their limitations, and how our work addresses and mitigates those limitations.
\subsection{Bias}

The term \textit{bias} is defined and used in many ways, as shown in \citet{olteanu2019}. The normative definition of bias 
in cognitive science, is: \textit{``Behaving according to some cognitive priors and presumed realities that might not be true at all''}~\cite{munoz2021}. 
Furthermore, the statistical definition of bias is \textit{``Systematic distortion in the sampled data that compromises its representatives''}~\cite{olteanu2019}. 
In NLP, while bias has been described in several ways, the statistical definition is the most used in the literature \cite{Caliskan2017, Garg2017, nangia-etal-2020-crows, nadeem-etal-2021-stereoset}. In this paper, similar to other works on investigating bias in the literature, we use the statistical definition of bias to measure the socio-technical bias.

Furthermore, in the last few years, various metrics have been proposed in the literature to quantify bias in static word embeddings \cite{Caliskan2017, Garg2017, sweeney2019, dev2019, elsafoury_sos_2022} and contextual word embeddings (language models) \cite{DBLP:conf/naacl/MayWBBR19, kurita-etal-2019-measuring, nangia-etal-2020-crows, nadeem-etal-2021-stereoset,Guo-and-caliskan-2021-ceat}. Other researchers focused on quantifying the NLP models' fairness when used in a downstream task \cite{Dearteaga2019, Borkan-etal-2019-naunced-metrics, qian-etal-2022-perturbation}, with most of these studies focusing on measuring only representation bias in LMs using various proposed metrics, such as CrowS-Pairs \cite{nangia-etal-2020-crows}, StereoSet \cite{nadeem-etal-2021-stereoset}, and SEAT \cite{DBLP:conf/naacl/MayWBBR19}.

\subsection{Fairness}
\textit{Fairness} has many formal definitions, built on those from literature on the fairness of exam testing from the 1960s, 70s and 80s \cite{Hutchinson-and-Mitchel-2019-50-years}.  The most recent fairness definitions are broadly categorized into two groups: 
\begin{itemize}
    \item \textit{Individual fairness}, which is defined as \say{\textit{An algorithm is fair if it gives similar predictions to similar individuals}}~\cite{kusner2017counterfactual}. Counterfactual fairness is an example of individual fairness metrics and there are different proposed metrics in the literature to measure counterfactual fairness \cite{kusner2017counterfactual, qian-etal-2022-perturbation, Krishna2022MeasuringFO,fryer-etal-2022-flexible}. 

    For a given model $\hat{Y}: X \rightarrow Y$ with features $X$, sensitive attributes $A$, prediction $\hat{Y}$, and two individuals $i$ and $j$, and if individuals i and j are similar, the model achieves individual fairness if $ \hat{Y}(X^i, A^i) \approx \hat{Y}(X^j, A^j) $.

    \item \textit{Group fairness}, which can be defined as \textit{``An algorithm is fair if the model prediction $\hat{Y}$ and sensitive attribute $A$ are independent''}~\cite{caton2020fairness, kusner2017counterfactual}. Based on group fairness, the model is fair if $\hat{Y}(X| A=0) = \hat{Y}(X|A=1)$. Group fairness is the most common definition used in NLP and there are two main approaches for measuring a model's fairness in that case: 
\begin{enumerate}

    \item \textit{Threshold-based metrics}, where a model's fairness is measured by how much the classifier's predicted labels differ between different groups of people, based on a threshold \cite{Hutchinson-and-Mitchel-2019-50-years}. The equalized odds metric is a threshold-based metric, and it is the most commonly used metric in the literature for measuring fairness in the downstream task of text classification \cite{Dearteaga2019, cao2022, steed-etal-2022-upstream}. Equalized odds are measured by the absolute difference ($gap$) between the true positive rates (TPR) or false positive rates (FPR) between different groups of people, $g$ and $\hat{g}$, based on sensitive attributes like gender, race, etc, as defined in Equations \ref{eq:fpr-gap} and \ref{eq:tpr-gap}.
    
    \begin{equation}
    FPR\_gap_{g, \hat{g}} = |FPR_{g} - FPR_{\hat{g}}|  
    \label{eq:fpr-gap}
    \end{equation}
    \begin{equation}
    TPR\_gap_{g, \hat{g}} = |TPR_{g} - TPR_{\hat{g}}|  
    \label{eq:tpr-gap}
    \end{equation}
    
    \item \textit{Threshold-agnostic metrics}, where fairness is measured by how much the distribution of the classifier's prediction probabilities varies across different groups of people based on their sensitive attributes. \citet{Borkan-etal-2019-naunced-metrics} propose a set of fairness metrics that are based on the AUC score. An important advantage of the threshold-based metrics is that they are robust to data imbalances in the amount of positive and negative examples in the test set \cite{Borkan-etal-2019-naunced-metrics}. In this work, we measure the absolute difference in the area under the curve (AUC) scores between marginalized group ($g$) and non-marginalized group $\hat{g}$, as shown in Equation~\ref{eq:auc-gap}. 
    \begin{equation}
    AUC\_gap_{g, \hat{g}} = |AUC_{g} - AUC_{\hat{g}}|  
    \label{eq:auc-gap}
\end{equation}
\end{enumerate}

\end{itemize}
\noindent
Nevertheless, the current metrics used to measure bias and fairness in NLP have some limitations.
For example, there is a lack of articulation of what the different metrics actually measure, and ambiguities and unstated assumptions, as discussed in \citet{Blodgett-etal-2021-norweigan-salmon}. There are also limitations with using sentence templates to measure the bias and fairness of LMs on downstream tasks. For example, \citet{seshadri2022quantifying} demonstrate that fairness scores vary with small lexical changes in the sentence templates that do not change the semantics of the sentences. However, these metrics could still be used as initial indicators of the Socio-technical bias in LMs and downstream NLP tasks. In this work, we use these metrics in that capacity as initial indicators of bias in LMs and of how they impact the fairness of toxicity detection. To mitigate for these limitations, we start our investigation using different group fairness metrics. Then, we use individual fairness metrics to further confirm the results of the group fairness metrics.

Furthermore, the focus of studying bias in the literature has been on LMs trained on English data and from a Western perspective where the marginalized identities tend to be women, non-white ethnicities, and minority religious groups like Muslims and Jews. Similarly, due to the datasets used in our paper, we focus our analysis on LMs trained on English data, and we investigate bias from a Western-perspective and focus our analysis on three sensitive attributes: Gender, Race, and Religion. 

\subsection{Related work}
The impact of bias on models' fairness on NLP downstream tasks is understudied. The majority of studies have focused on the impact of representation bias \cite{cao2022, kaneko2022, goldfarb-tarrant-etal-2021-intrinsic}, excluding the impact of other sources of bias like selection and overamplification. 
Some researchers found no strong evidence that representation bias in LM impacts fairness in downstream NLP tasks \cite{cao2022, kaneko2022}. \citet{steed-etal-2022-upstream} found a positive correlation between representation bias and fairness of downstream tasks but argued that this positive correlation is a result of cultural artifacts found in both pre-training and fine-tuning datasets. 

However, there are some limitations to those studies. For example, in \citet{steed-etal-2022-upstream}, the authors use two representation bias metrics which use bleached template sentences, which are sentences that don't have a real semantic context, to measure bias, these metrics have been criticized as they may not be semantically bleached \cite{DBLP:conf/naacl/MayWBBR19}. Since the used bias metrics are not reliable, their impact on the fairness of NLP tasks is also unreliable. Moreover, both \citet{cao2022} and \citet{steed-etal-2022-upstream} use different representation bias metrics for the two text classification tasks examined, which results in a lack of consistency in their findings.

Similarly, the impact of removing bias on the models' fairness in downstream tasks has been understudied. For example, \citet{meade-etal-2022-emprical-survey-debias} investigate the impact that different debiasing approaches have on the performance of different NLP downstream tasks. However, they don't investigate the impact debiasing has on the fairness of the downstream tasks. \citet{baldini-etal-2022-fairness} investigate the impact of removing bias from fine-tuned models, excluding the impact of removing representation bias on the fairness of downstream tasks.

On the other hand, in \citet{kaneko2022}, the authors investigate the effectiveness of different debiasing methods that remove different sources of bias representation bias and overamplification bias on the fairness of the downstream tasks. However, the authors do not investigate the impact of removing selection bias on the fairness of downstream NLP tasks. Moreover, for removing overamplification bias, the authors' investigation covers only gender bias in the downstream tasks of occupation classification, semantic textual similarities (STS) and natural language inference (NLI), excluding other bias types like racial and religion bias and tasks like toxicity detection. 

Similarly, \citet{steed-etal-2022-upstream} investigate the impact of different bias removal techniques to remove representation and selection bias on the fairness of downstream NLP tasks. However, the authors don't investigate the impact of removing overamplification bias on the fairness of downstream NLP tasks. Additionally, the authors apply the selection bias removal investigation only on gender bias and only for the task of occupation classification.

As for measuring models' fairness on downstream tasks, \citet{goldfarb-tarrant-etal-2021-intrinsic} measure fairness as performance gap between different identity groups. However, more recently, fairness is measured as the gap in the false or true positive rates between the different identity groups \cite{Dearteaga2019, steed-etal-2022-upstream, baldini-etal-2022-fairness, kaneko2022}.  

On the other hand, \citet{cao2022, kaneko2022, steed-etal-2022-upstream,baldini-etal-2022-fairness}, use only threshold-based fairness bias metrics for the text classification task. For example, \citet{cao2022,kaneko2022} use FPR gap to measure fairness on toxicity detection. Similarly, \citet{steed-etal-2022-upstream} use TPR gap to measure fairness in the task of occupation classification and FPR gap for the task of toxicity detection. However, more recent research demonstrate that different group fairness metrics give different results \cite{jourdan-etal-2023-fairness} .

Threshold-agnostic metrics have not been widely used to measure fairness and to investigate its correlation to representation bias. Even though, according to \citet{Borkan-etal-2019-naunced-metrics}, threshold-agnostic metrics can capture the behavior of the model. Similarly, individual fairness metrics, e.g., counterfactual fairness, have not been used in investigating the impact of bias. Even though, counterfactual fairness provides an in-depth analysis of how the model treats different sentences based on the identity group present in the sentences. Moreover, in most of the studies that investigate the fairness of the task of toxicity detection, the authors don't explain how they measure the fairness of the models between the different identity groups \cite{cao2022,steed-etal-2022-upstream}. Other researchers do not measure fairness as a notion of discrimination between marginalized and non- marginalized groups but rather as the mention of sensitive topics. For example, \citet{baldini-etal-2022-fairness} measure fairness using ''\textit{coarse-grained groups (e.g., mention of any
religion) instead of the fine-grained annotations (e.g., Muslim)}''. Since the authors do not actually measure fairness but rather how a model treats the existence of a sensitive topic, the findings of their work might not be reliable or extendable to fairness as a notion of discrimination between different identity groups.

This paper fills these research gaps in the literature by investigating different sources of bias and their impact on the models' fairness in the downstream task of toxicity detection. We investigate bias and fairness for different sensitive attributes: gender, race, and religion. We aim to overcome the limitations of previous research by using different metrics to measure representation bias and models' fairness. Moreover, we investigate the effectiveness of various bias mitigation methods (debiasing) for removing different sources of bias: representation, selection and overamplification, as well as their impact on the fairness and performance of the task of toxicity detection.

\section{Methodology}
\begin{figure}
\centering
    \includegraphics[width=0.6\textwidth]{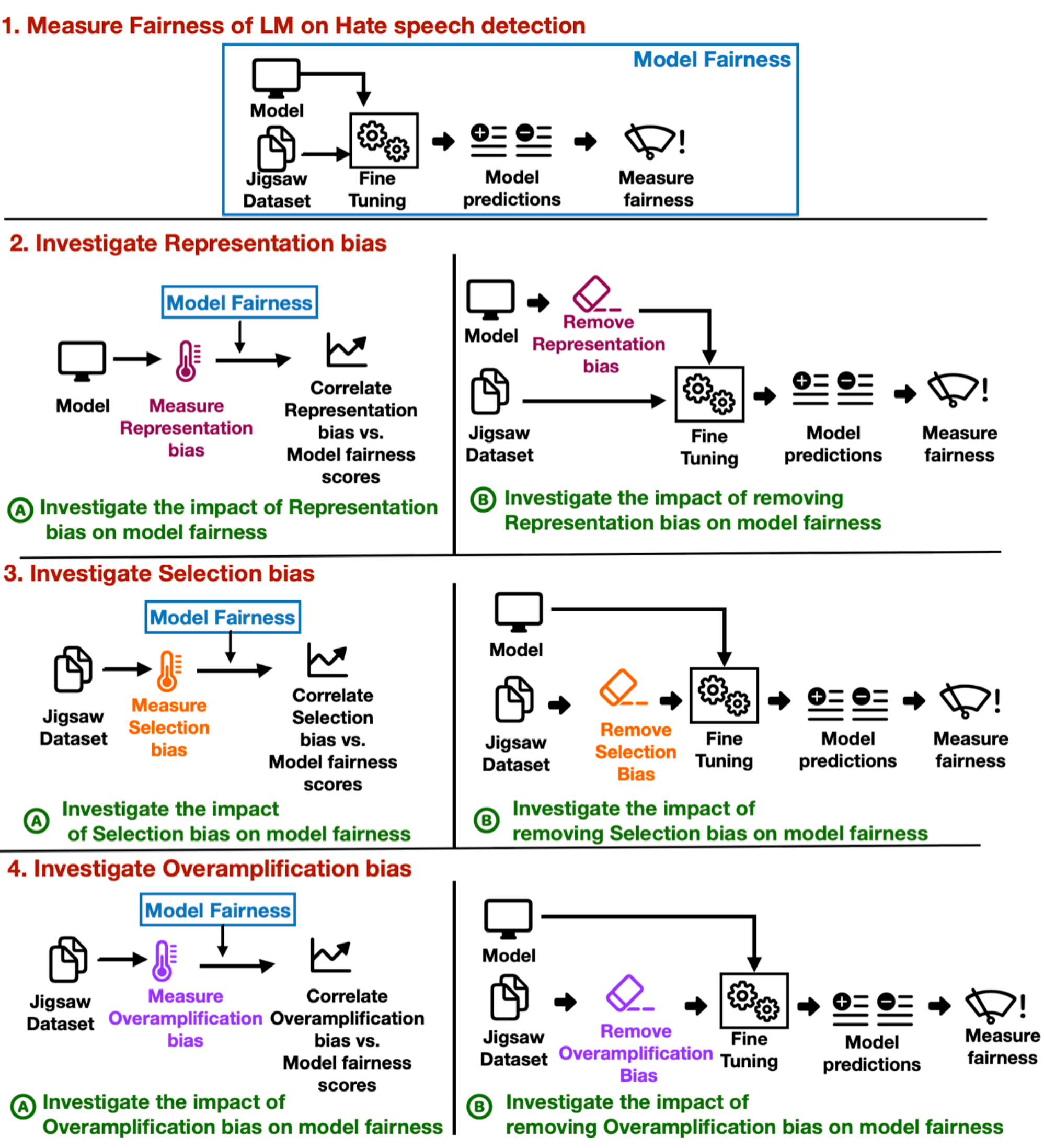}
    \caption{Overview of conducted investigation.}
    \label{fig:paper_overview}
\end{figure}

Four groups of experiments were conducted to investigate the impact of each source of bias on the fairness of the downstream task of toxicity detection. 
\figurename~\ref{fig:paper_overview} provides an overview of these four groups of experiments. As shown in \figurename~\ref{fig:paper_overview},  in {Step 1}, we first measured the fairness of the toxicity detection task (\cref{sec:extrinsic_bias}) using various models and used these fairness scores as a baseline. Then, in {Step 2A}, we measured the representation bias in the inspected models and its impact on the models' fairness on the task of toxicity detection (\cref{sec:intrinsic_bias}), as well as the impact of removing representation bias (\cref{sec:sentDebias}), as shown in {Step 2B} of \figurename~\ref{fig:paper_overview}. In {Steps 3 (A \& B) and 4 (A \& B)}, we repeated the same investigation for selection bias (\cref{sec:selection_bias} \& \cref{sec:selection-bias-removal}) and overamplification bias (\cref{sec:overap_bias} \& \cref{sec:over-bias-removal}). Then, we investigated the impact of removing multiple sources of biases on the fairness of the task of toxicity detection (\cref{sec:multibiases}). 
Finally, we build on these findings and recommend guidelines to achieve fairer toxicity detection (\cref{sec:improving_fainress}).

\section{Toxicity detection}
\label{sec:toxicity_detection}
\subsection{Dataset}
We use the civil community dataset \cite{Borkan-etal-2019-naunced-metrics} in our experiments. The dataset contains almost 2 million comments from the Civil Comments Platform, labelled as toxic or not, along with labels on the identity of the target of the sentence, e.g., religion, sexual orientation, gender, and race. The identity labels provided in the dataset are both crowd-sourced and automatically labelled. When we analyzed the dataset, we found some issues with the identity labels, e.g., some data items are labelled to contain more than one identity (male and female) as the target of the toxicity\footnote{\url{https://huggingface.co/datasets/google/civil_comments}}. 
Then, the dataset was pre-processed to keep only the data items where the identity information is labelled by human annotators. Additionally, we follow the same data pre-processing steps used in \citet{elsafoury2021}, where the authors train a BERT model for the task of cyberbullying detection. To this end, we remove URLs and non-ASCII characters, lowercase all the letters, convert all contractions to their formal format and add a space between words and punctuation marks. This resulted in 400K data items. The dataset is then split into 40\% training, 30\% validation, and 30\% test sets. We limit this investigation to 3 sensitive attributes, i.e., gender, religion, and race, as shown in \tablename~\ref{tab:senstive_attributs_used_in_ch6}.

 \begin{table}
     \centering
     \begin{tabular}{l|l|l} \hline
         Sensitive attribute & Marginalized & Non-marginalized \\ \hline
         Gender & Female & Male \\ \hline
         Race & Black, Asian & White \\ \hline
         Religion &Jewish, Muslim & Christian\\ \hline
     \end{tabular}
     \caption{The examined sensitive attributes and identity groups.}
     \label{tab:senstive_attributs_used_in_ch6}
 \end{table}

We only use the civil community dataset because, to the best of our knowledge, it is the only available toxicity dataset that contains information on both marginalized and non-marginalized identities, which is important to the way we measure fairness, as explained in section~\ref{sec:extrinsic_bias}. Other datasets, like ToxiGen \cite{hartvigsen-etal-2022-toxigen}, SocialFrame \cite{sap-etal-2020-social}, Ethos \cite{mollas_ethos_2022}, and the MLM data \cite{ousidhoum-etal-multilingual-hate-speech-2019} contain information only about marginalized groups, and thus cannot be used in our investigation. The HateXplain dataset \cite{mathew2021hatexplain} contains information about both marginalized and non-marginalized identities. However, HateXplain uses offensive words to refer to marginalized groups, e.g., n*gger to refer to Black people, and identity words to describe non-marginalized groups, e.g., white to refer to Caucasians. This makes the HateXplain dataset unsuitable for the experiments held in section~\ref{sec:extrinsic_bias} where we create data perturbations, since replacing an offensive word that describes marginalized groups with an identity word to describe a non-marginalized identity group changes the meaning of the sentence and hence its label as hateful or not. Conversely, in the civil community dataset, identity words are used to describe both marginalized and non-marginalized groups. Even the toxic sentences in the civil community dataset do not contain offensive words to describe marginalized groups. This is evident in the most common nouns and adjectives found in the toxic and non-toxic sentences that are targeted at the different identity groups, as shown in \tablename~\ref{tab:most-common-words}. It must be noted that the most common nouns and adjectives were generated using the Spacy Python package\footnote{\url{https://spacy.io/api}}.

\begin{table}[t]
    \centering
    \begin{tabular}{l|l|l} \hline
    Identity & Toxic sentences & Non-toxic sentences  \\ \hline
    Black & \makecell{black, people, blacks, racist,\\ police,Black, other,\\ man, white, men} 
    & \makecell{black, people, blacks, man, \\police,other, Black, \\white, many, men} \\ \hline
    
    Asian & \makecell{people, Asian, many, repair,\\ chef, country, racist, \\real, citizens, Korean} &
    \makecell{Asian, other, Chinese, people,\\ many, countries, years,\\ women, more, country} \\ \hline
    
    White & \makecell{white, people, racist, men, \\supremacists, man, racism,\\ right, supremacist, White} &
    \makecell{white, people, men, racist,\\ right, other, man,\\ many, supremacists, male} \\ \hline
   
    Female & \makecell{women, woman, people, white,\\ other, many, sexual,\\ time, life, sex} &
    \makecell{women, woman, people, many, \\ other, more, time, \\ right, life, abortion} \\ \hline
    
     Male & \makecell{man, men, white, black,\\ people, male, women, \\ stupid, racist, males} & 
    \makecell{man, men, white, people,\\ male, other, many,\\ right, time, way} \\ \hline

    Muslim & \makecell{Muslim, people, women, white,\\ many, other, muslim,\\ terrorists, religion, muslims} &
    \makecell{Muslim, people, countries, women,\\ other,
    many, country,\\ ban, world, muslim} \\ \hline

    Jewish & \makecell{Jewish, people, anti, black,\\
    hate, women, good,\\ other, white, man} &
    \makecell{Jewish, people, anti, other,\\
    white, right, way,\\ state, many, world} \\ \hline

    Christian & \makecell{people, white, Christian, women,\\
    right, other, many,\\ sex, Catholic,life} &
    \makecell{Catholic, people, Christian, \\ church, many,
    women, \\ other, right, time, good} \\ \hline
    \end{tabular}
    \caption{The most common ten adjectives and nouns as generated by the Spacy Python package in the toxic and non-toxic sentences that are targeted at different identity groups in the civil community fairness dataset.}
    \label{tab:most-common-words}
\end{table}

\subsection{Language models}
The fairness of the downstream task of toxicity detection is evaluated on the following widely used models: BERT-base-uncased \cite{DBLP:conf/naacl/DevlinCLT19}, RoBERTa-base \cite{roberta}, and ALBERT-base \cite{albert}. We fine-tune these models on the civil community dataset. Following the experimental setting from \cite{elsafoury2021}, the models are fine-tuned for 3 epochs, using a batch size of 32, a learning rate of $2e^{-5}$, and a maximum text length of 61.  
Classification results using the fine-tuned models indicate that ALBERT-base is the best-performing model, with an AUC score of 0.911, followed by RoBERTa-base with an AUC score of 0.908, and BERT-base with an AUC score of 0.902. The fine-tuned models are then used to measure fairness in the toxicity detection task.

\section{Fairness in the task of toxicity detection}
\label{sec:extrinsic_bias}

To evaluate the fairness of the examined models on the downstream task of toxicity detection, we used two sets of fairness metrics: (i) Threshold-based, which uses the absolute difference ($gap$) in the false positive rates ($FPR$) and true positive rates ($TPR$) between the marginalized group ($g$) and non-marginalized group $\hat{g}$, as shown in Equations \ref{eq:fpr-gap} and \ref{eq:tpr-gap}, and (ii) Threshold-agnostic metrics, which measure the absolute difference in the area under the curve (AUC) scores between marginalized group ($g$) and non-marginalized group $\hat{g}$, as shown in Equation \ref{eq:auc-gap}. In cases where there is more than one identity group in the marginalized group for a sensitive attribute, e.g., Asian and Black vs. White, we measured the mean of the FPR, TPR, and AUC scores of the two groups, e.g., Asian and Black, and then use that score to represent the marginalized group ($g$). These scores express the amount of unfairness in the toxicity detection models, with higher scores denoting less fair models and lower scores denoting fairer models. These fairness metrics are measured between two groups, marginalized and non-marginalized, similar to the approach used by \citet{elsafoury_sos_2022}. 

\subsection{Balanced fairness dataset}
\label{sec:balanced_toxiciy_fairness_dataset}

To measure fairness, we filtered the test set to ensure that the sentences contain only one identity group, which resulted in 21K sentences to improve the quality of the measured fairness. We found differences in the sizes of the subsets of sentences that mention the different identity groups. For example, the size of the subset of sentences that are targeted at males is 3,716 sentences while the size of the female subset is 6,046 sentences. We also found differences in the ratio of the positive samples between the different identity groups that belong to the same sensitive attribute. The ratio of positive samples for the male and female groups are 0.12 and 0.10 respectively; for the White, Asian, and Black groups are 0.20, 0.07, and 0.27 respectively; and for the Christian, Muslim, and Jewish groups are 0.05, 0.16, and 0.12 respectively. We hypothesize that these differences between the different identity groups may influence the fairness scores.

To test this hypothesis, we created a balanced fairness dataset and used it to measure fairness in the fine-tuned models. To create this balanced fairness dataset, we followed previous research and augmented our dataset using perturbations \cite{Webster-etal-2020-cda}. Previous studies, like \citet{Sahaj2019} and \cite{Fryer2022}, used data perturbations to improve the fairness of the task of toxicity detection. However, the authors of these two studies chose to create the data perturbations only on the non-toxic sentences to avoid what \citet{Sahaj2019} describe as \textit{Asymmetric Counterfactuals}. \citet{Sahaj2019} argue that asymmetric counterfactuals happen in toxic sentences for two reasons: 
\begin{enumerate}
    \item When the toxicity targets a marginalized group, it is based only on the identity, and no other toxicity signals are present. In such cases substituting identities would result in non-toxic sentence and hence equal prediction should not be required over most counterfactuals.

    \item Stereotyping comments are more likely to occur in a toxic comment attacking the marginalized group than in a non-toxic comment referencing some other identity group.
\end{enumerate}

However, the results in \tablename~\ref{tab:most-common-words} show that the most common nouns and adjectives in the toxic and non-toxic sentences that target marginalized and non-marginalized groups are almost similar. For example, words like \say{black}, \say{blacks}, \say{police} are used in both toxic and non-toxic comments to describe Black people and words like \say{white}, \say{racist}, \say{supremacists} exist in both toxic and non-toxic sentences that describe White people. This challenges the hypothesis made by \citet{Sahaj2019} that stereotypes are more likely to happen in toxic sentences. Additionally, the offensive words found in the toxic comments targeted at marginalized groups are not based on identity. For example, \tablename~\ref{tab:most-common-words} shows that the word \say{racist} is used in the toxic comments targeted at White, Black, and Asian people. This challenges the second hypothesis made by \citet{Sahaj2019} that toxicity targeted at marginalized identities is based mainly on the identity with no other toxicity signal. These findings suggest that in our dataset, asymmetric counterfactuals are not more likely to happen in toxic sentences than non-toxic sentences. Hence and based on this evidence, in this work, we decided to create perturbations for both toxic and non-toxic sentences.

To this end, lexical word replacement was used to create perturbations of existing sentences using regular expressions. As explained earlier, this is possible with the civil community dataset because after inspecting the most common nouns and adjectives used in each subset that targets a certain identity, it was found that the most common words are non-offensive words that describe that identity. For example, among the most common nouns in the data subset that are targeted at black people are: \say{black} and \say{blacks}. The most common nouns in the data subset targeted at Asian people are \say{asian} and \say{chinese}. A similar pattern is found for religion and gender identities. However, this approach is not suitable for gender perturbations, as, in the English language, pronouns also change between males and females. To this end, perturbations for the male and female identity groups are created using the AugLy\footnote{\url{https://augly.readthedocs.io/en/latest/README.html}} tool, which is provided by Facebook research to swap gender information~\cite{papakipos2022augly}.

The balanced fairness dataset contains 55,476 samples and has the same ratio between positive and negative samples for each identity group within the same sensitive attribute. For example, the ratio of the positive (toxic) examples in the male and female identity groups is 0.10 for the gender attribute, the ratio of the positive samples for the Black, White and Asian groups is 0.20 for the race attribute, and the ratio of the positive samples for the Muslim, Christian and Jewish groups is 0.10 for the religion attribute.

\subsection{Fairness results}
The fairness evaluation results for the three examined models on the original and the balanced civil community fairness datasets are shown in \tablename~\ref{tab:extrinsic_bias_scores_perturbed_vs_original_jigsaw_dataset}. From this table, it is evident that the use of the balanced civil community fairness dataset led to improved fairness scores across most of the fairness metrics and across all models. This finding suggests that the dataset used to measure the fairness in the downstream task of toxicity detection impacts the measured fairness, and it is important to ensure that there is a balanced representation of the different identity groups to get reliable fairness scores. This finding supports our hypothesis and, to the best of our knowledge, has not been mentioned in the literature on measuring fairness before. 

The results also indicate that when we use the original imbalanced fairness dataset, the different metrics used to measure the fairness reported different fairness scores related to each sensitive attribute in the fine-tuned models. We used Pearson's correlation coefficient ($\rho$) to measure how different the fairness metrics are and found that even though the FPR\_gap and TPR\_gap are both threshold-based metrics, there is no positive correlation between the two metrics for the three models. There is a negative correlation between TPR\_gap and FPR\_gap ($\rho$ = -0.37), a negative correlation between the TPR\_gap and the AUC\_gap scores for the three models ($\rho$ = -0.42), and a positive correlation between the FPR\_gap and the AUC\_gap for the models ($\rho$ = 0.46).

On the other hand, when we used the balanced civil community fairness dataset to measure fairness, we found a positive correlation between all the fairness metrics in all three models. There is a positive correlation between the FPR\_gap and the TPR\_gap scores ($\rho$ = 0.59), a positive correlation between FPR\_gap and AUC\_gap scores ($\rho$ = 0.64), and a positive correlation between the TPR\_gap and the AUC\_gap scores ($\rho$ = 0.27). This is another evidence that using a fairness dataset with a balanced representation of the different identity groups leads to more reliable fairness scores. For the rest of this work, the balanced fairness dataset is used to measure fairness. Similarly, in the rest of the paper, to investigate the impact of the different sources of bias in the next sections, we use Pearson's correlation between the bias scores and the fairness scores measured in this section similar to the work done in \citet{steed-etal-2022-upstream, kaneko2022, cao2022}.

\begin{table}
    \centering
    \renewcommand{\arraystretch}{1.10}
    \resizebox{0.8\textwidth}{!}{
    \begin{tabular}{lllrrr}
    \hline
         Attribute & Model & Dataset  & FPR\_gap & TPR\_gap & AUC\_gap  \\
    \hline
         \multirow{6}{*}{Gender} &
         \multirow{2}{*}{ALBERT} &
         Original & 0.001 & 0.081 & 0.025 \\
    & &  Balanced  & \color{red}$\uparrow0.006$    &  \color{teal}$\downarrow0.038$   &  \color{teal}$\downarrow0.003$ \\
    \cline{2-6}
    &     \multirow{2}{*}{BERT}
      &  Original  & 0.002 &  0.111 & 0.026 \\
    & &  Balanced  & \color{red}$\uparrow0.008$    &  \color{teal}$\downarrow0.036$   &  \color{teal}$\downarrow0.009$ \\
    \cline{2-6}
    &     \multirow{2}{*}{RoBERTa}
      &  Original  & 0.007 &  0.084 & 0.017 \\
    & &  Balanced  & \color{teal}$\downarrow0.004$    &  \color{teal}$\downarrow0.031$   &  \color{teal}$\downarrow0.011$ \\
    \hline
         \multirow{6}{*}{Race} &
         \multirow{2}{*}{ALBERT} &
         Original &  0.007 &   0.044 &  0.003\\
    & &  Balanced &  \color{red}$\uparrow0.008$   &   \color{teal}$\downarrow0.0016$   &  \color{red}$\uparrow0.018$ \\
    \cline{2-6}
    &     \multirow{2}{*}{BERT}
      &  Original &  0.008 &   0.017 &  0.048 \\
    & &  Balanced &  \color{red}$\uparrow0.015$   &   \color{teal}$\downarrow0.002$   &  \color{teal}$\downarrow0.025$ \\
    \cline{2-6}
    &     \multirow{2}{*}{RoBERTa}
      &  Original  &  0.014 &   0.127 &  0.028 \\
    & &  Balanced  &  \color{teal}$\downarrow0.003$   &   \color{teal}$\downarrow0.011$  &  \color{teal}$\downarrow0.021$ \\
    \hline
         \multirow{6}{*}{Religion} &
         \multirow{2}{*}{ALBERT} &
         Original   &  0.019 &  0.060  &   0.042 \\
    & &  Balanced  &   \color{teal}$\downarrow0.009$  &   \color{red}$\uparrow0.108$   &  \color{teal}$\downarrow0.020$ \\
    \cline{2-6}
    &     \multirow{2}{*}{BERT}
      &  Original   &  0.016 &  0.027  &   0.051 \\
    & &  Balanced &   \color{teal}$\downarrow0.008$  &   \color{red}$\uparrow0.062$   &  \color{teal}$\downarrow0.012$ \\
    \cline{2-6}
    &     \multirow{2}{*}{RoBERTa}
      &  Original   &  0.027 &  0.030  &   0.0369 \\
    & &  Balanced  &   \color{teal}$\downarrow0.021$  &   \color{red}$\uparrow0.160$   &  \color{teal}$\downarrow0.027$ \\
    \hline
    \end{tabular}
    } 
    \caption{The fairness scores of the examined models on the original and the balanced civil community fairness datasets. (\textcolor{red}{$\uparrow$}) denotes that the fairness score increased, and the fairness worsened. (\textcolor{teal}{$\downarrow$}) denotes that the fairness score decreased, and the fairness improved.}
    \label{tab:extrinsic_bias_scores_perturbed_vs_original_jigsaw_dataset}
\end{table}

\begin{figure}[]
    \centering
    \includegraphics[width=0.7\columnwidth]{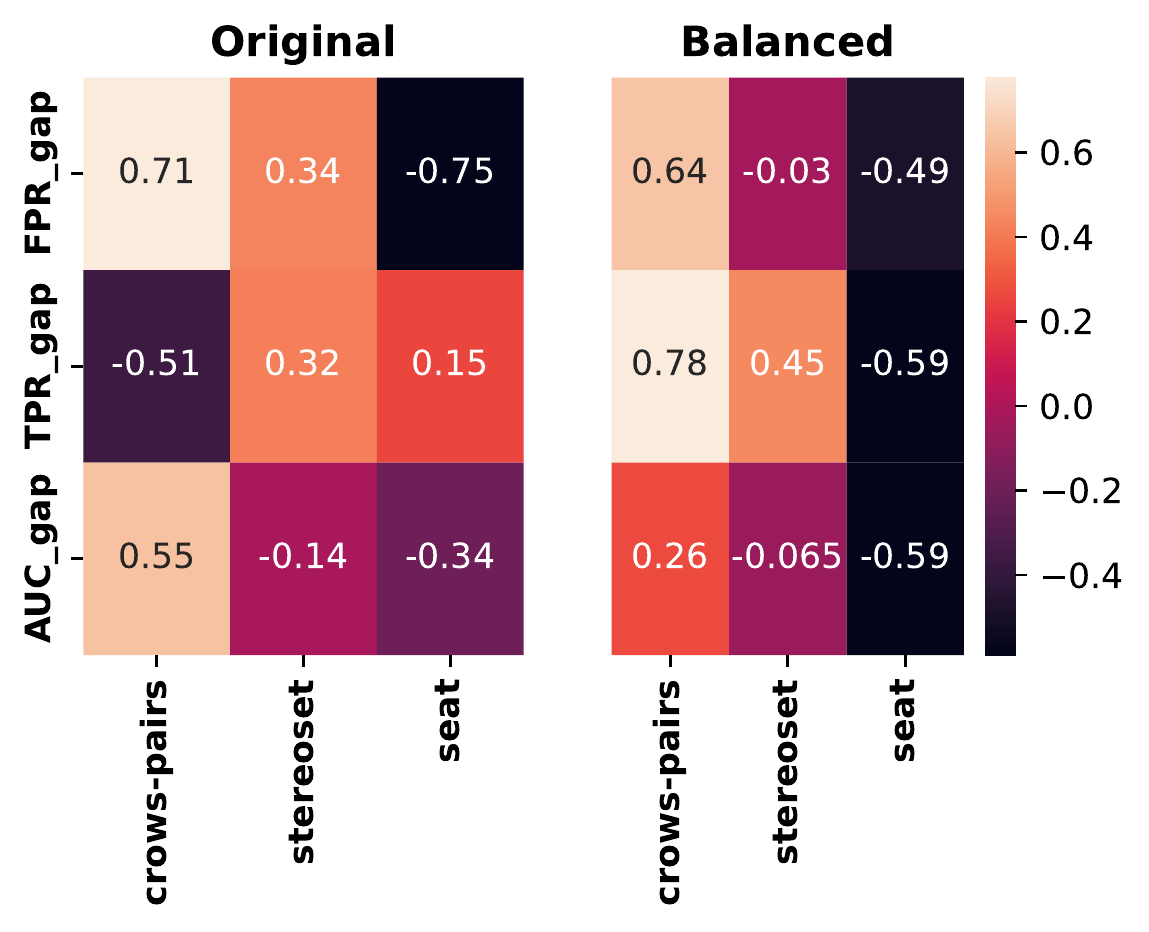}
    \caption{Heatmap of Pearson's correlation between representation bias scores of all LMs and fairness scores of LMs on the downstream task of toxicity detection, on the original civil community fairness dataset (left) and the balanced civil community fairness dataset (right), for all the sensitive attributes.}
    \label{fig:intrinsic_vs_extrinsic_bias}
\end{figure}
\section{Sources of bias}
\citet{shah-etal-2020-predictive} present a framework for bias in NLP which includes four sources of bias in NLP models that impact a model's outcome, which are representation bias, label bias, selection bias, and overamplification bias. In this work, we examine how the following three sources of bias impact the examined models' fairness on the task of toxicity detection: (i) representation bias, (ii) selection bias (also called sample bias), and (iii) overamplification bias. We do not investigate label bias because there is no information available on the annotators of the examined civil community dataset. Furthermore, we removed each source of bias and investigated the impact of the bias removal on the fairness of toxicity detection.

\subsection{Representation bias}
\label{sec:intrinsic_bias}
Representation bias describes the societal stereotypes that language models encoded during pre-training. We used three metrics to measure representation bias: CrowS-Pairs \cite{nangia-etal-2020-crows}, StereoSet \cite{nadeem-etal-2021-stereoset}, and SEAT \cite{DBLP:conf/naacl/MayWBBR19} to measure three types of social bias: gender, religion, and race. As shown in \tablename~\ref{tab:intrinsic_bias_scores}, the results indicate that RoBERTa is the most biased model according to CrowS-Pairs and StereoSet.

We then investigated the impact of representation bias in the inspected models, BERT, ALBERT, and RoBERTa, on their fairness on the task of toxicity detection. To measure that impact, we measure the Pearson's correlation coefficient ($\rho$) between fairness scores measured by the different fairness metrics and the representation bias scores measured by the different representation bias metrics (\figurename~\ref{fig:intrinsic_vs_extrinsic_bias} (right)). We found a consistent positive correlation between the CrowS-Pairs bias scores with the fairness scores measured by all three fairness metrics (FPR\_gap. TPR\_gap and AUC\_gap) for all the models and sensitive attributes. On the other hand, there is an inconsistent correlation with the StereoSet scores. This finding is different from previous research that suggested that there is no correlation between representation bias and fairness scores \cite{cao2022, kaneko2022}. We hypothesize that previous research did not use a balanced fairness dataset, which is why they did not find a consistent positive correlation with fairness metrics. To test this hypothesis, we measured the Pearson correlation coefficient ($\rho$) between representation bias scores and fairness scores measured using the different fairness metrics on the original imbalanced fairness dataset. We found no consistent positive correlation between representation bias and fairness scores, which supports our hypothesis as shown in \figurename~\ref{fig:intrinsic_vs_extrinsic_bias} (left). Another reason why previous research has not found a consistent positive correlation is that they used only one metric for either representation bias or fairness. However, using more than one metric helped to reveal this correlation.

\begin{table}
    \renewcommand{\arraystretch}{1.3}
    \resizebox{1\textwidth}{!}{
\begin{tabular}{l|r|r|r|r|r|r|r|r|r}
\hline
Model & \multicolumn{3}{c|}{CrowsPairs} & \multicolumn{3}{c|}{StereoSet} & \multicolumn{3}{c} {SEAT}\\ \hline

                              & Gender    & Race      & Religion  
                              & Gender   & Race     & Religion   
                              & Gender   & Race    & Religion 
                               \\\hline
\textbf{AlBERT-base}          & 0.541    & 0.513     & 0.590 
& 0.599    & 0.575     & 0.603
& 0.622    & 0.551    & 0.430   
   \\ \hline
+ SentDebias-gender           & \color{teal}$\downarrow0.461$    & \color{teal}$\downarrow0.436$      & \color{teal}$\downarrow0.466$     

& \color{teal}$\downarrow0.517$    & \color{teal}$\downarrow0.552$ & \color{teal}$\downarrow0.586$     
& 0.622    & 0.551    & 0.430   
  \\ \hline
  
+ SentDebias-race             & \color{red}$\uparrow0.564$    & \color{teal}$\downarrow0.440$     & \color{red}$\uparrow0.666$   
& \color{teal}$\downarrow0.542$    & \color{teal}$\downarrow0.521$     & \color{teal}$\downarrow0.555$   
& 0.622    & 0.551    & 0.430   
    \\ \hline

+ SentDebias-religion         & \color{red}$\uparrow0.549$    & \color{red}$\uparrow0.660$     & \color{teal}$\downarrow0.581$  
& \color{teal}$\downarrow0.501$    & \color{teal}$\downarrow0.529$     & \color{teal}$\downarrow0.510$  
& 0.622    & 0.551    & 0.430   
   \\ \hline


\textbf{BERT-base-uncased}    & 0.580    & 0.581      & 0.714  
& 0.607    & 0.5702     & 0.597    
& 0.620    & 0.620    & 0.491  
   \\ \hline

+ SentDebias-gender           & \color{teal}$\downarrow0.427$    & \color{teal}$\downarrow0.555$      & \color{teal}$\downarrow0.647$    
& \color{teal}$\downarrow0.475$    & \color{teal}$\downarrow0.476$     & \color{teal}$\downarrow0.504$   
& 0.620    & 0.620    & 0.491 
      \\ \hline

+ SentDebias-race             & \color{teal}$\downarrow0.534$   & \color{teal}$\downarrow0.398$        &\color{teal}$\downarrow 0.704$   
& \color{teal}$\downarrow0.467$    & \color{teal}$\downarrow0.562$     & \color{teal}$\downarrow0.489$    
& 0.620    & 0.620    & 0.491 
    \\ \hline

+ SentDebias-religion         & \color{teal}$\downarrow0.534$   & \color{red}$\uparrow0.675$       & \color{teal}$\downarrow0.561$  
& \color{teal}$\downarrow0.469$    & \color{teal}$\downarrow0.511$     & \color{teal}$\downarrow0.399$   
& 0.620    & 0.620    & 0.491  
     \\ \hline


\textbf{RoBERTa-base}         & 0.606   & 0.527       & 0.771   
& 0.663     & 0.616      & 0.642   
& 0.939    & 0.307    & 0.126 
       \\ \hline

+ SentDebias-gender           & \color{teal}$\downarrow0.467$   & \color{red}$\uparrow0.691$       & \color{teal}$\downarrow0.561$ 
& \color{teal}$\downarrow0.518$    & \color{teal}$\downarrow0.497$      & \color{teal}$\downarrow0.477$   
& 0.939    & 0.307    & 0.126  

\\ \hline

+ SentDebias-race             & \color{teal}$\downarrow0.429$   & \color{teal}$\downarrow0.467$       & \color{teal}$\downarrow0.419$ 
& \color{teal}$\downarrow0.485$    & \color{teal}$\downarrow0.488$      & \color{teal}$\downarrow0.486$   
& 0.939    & 0.307    & 0.126   

\\ \hline

+ SentDebias-religion         & \color{teal}$\downarrow0.413$   & \color{teal}$\downarrow0.478$      & \color{teal}$\downarrow0.352$ 
& \color{teal}$\downarrow0.516$    & \color{teal}$\downarrow0.497$      & \color{teal}$\downarrow0.486$ 
& 0.939    & 0.307    & 0.126  
\\ \hline


\end{tabular}}
\caption{Representation bias scores in the examined models using different bias metrics before and after removing bias using the SentDebias algorithm. (\textcolor{red}{$\uparrow$}) denotes that the fairness metric score increased and the fairness worsened. (\textcolor{teal}{$\downarrow$}) denotes that the fairness metric score decreased, and the fairness improved.}
\label{tab:intrinsic_bias_scores}
\end{table}

\subsection{Representation bias removal}
\label{sec:sentDebias}
We used the SentDebias \cite{liang2020towards} to remove different types of bias from the LMs by projecting a sentence representation onto the estimated bias subspace and subtracting the resulting projection from the original sentence representation. The authors in \citet{liang2020towards} compute the bias subspace by following these steps:
\begin{enumerate}
    \item They define a list of identity words e.g., \say{woman/man}.
    \item They contextualize the identity words into sentences by finding sentences that contain those identity words in public datasets like SST \footnote{\url{https://huggingface.co/datasets/sst}} and WikiText-2 \footnote{\url{https://huggingface.co/datasets/mindchain/wikitext2}}.
    \item They obtain the representation of the contextualized sentence from the pre-trained LM.
    \item Principle Component Analysis (PCA) \cite{abdi2010principal} then used to estimate principal directions of variations of the sentences' representations. The first $K$ principal components are taken to define the bias subspace.
\end{enumerate}
We used SentDebias to remove gender, racial, and religious bias from the inspected models using the same code and experimental settings shared by \citet{meade-etal-2022-emprical-survey-debias}. SentDebias seems to reduce the bias scores, as shown in \tablename~\ref{tab:intrinsic_bias_scores}, in all the models according to the StereoSet and CrowS-Pairs metrics. On the other hand, the SEAT metric, did not show any difference in the bias scores for the debiased models, unlike the reported results in \citet{meade-etal-2022-emprical-survey-debias}. 
We found that, according to some metrics, removing one type of bias sometimes leads to exacerbating another type of bias. For example, in \tablename~\ref{tab:intrinsic_bias_scores}, removing racial bias increased the gender bias and removing religion bias increased racial and gender bias in AlBERT-base according to the CrowS-Pairs metric. The same finding is reported in \citet{elsafoury_sos_2022} for static word embeddings. 

Furthermore, we investigate the impact of removing representation bias on the models' fairness. We fine-tuned BERT-base, ALBERT-base and RoBERTa-base after debiasing them to remove gender, racial and religious bias. Then, we measure the fairness of the models using the different fairness metrics, threshold-based and threshold-agnostic metrics. The results in \tablename~\ref{tab:upstream_debias_fainress_jigsaw_dataset} indicate that, in some cases, removing representation bias, worsened the performance of the models slightly. In other cases, removing representation bias also improved the performance slightly. For example, removing gender bias information increased slightly the AUC scores, in BERT and RoBERTa. This is because the debiased models tend to predict more positive class examples, leading to more true positives and more false positives.

To simplify the analysis of the results, we investigated the impact of removing a certain type of bias on the fairness of the matching sensitive attribute. For example, we analyzed the impact of removing gender bias from the model representation (+ Upstream-SentDebias-gender) on the fairness of the models regarding the gender-sensitive attribute. For most of the models, the majority of the fairness metrics show that removing a certain type of bias from the model representation using SentDebias did not improve fairness for the corresponding sensitive attribute. There are improvements according to certain metrics, but these improvements are inconsistent across sensitive attributes and models. As for the cases where all fairness metrics show improvement in fairness, we find that only removing religion bias from RoBERTa-base representations improved the models' fairness for the religion sensitive attributes in the downstream task of toxicity detection. On the other hand, sometimes removing different types of bias from the model representation led to improvement in fairness for a different sensitive attribute. For example, in the BERT model, according to the FPR\_gap and TPR\_gap metric, removing racial bias information from the model's representation improved the fairness for the gender-sensitive attributes. Yet again, these findings are inconsistent across all models and across all fairness metrics.

These results suggest that even though there is a positive correlation between representation bias in the inspected models and the models' fairness on the task of toxicity detection bias scores, removing representation bias did not consistently improve the models' fairness. Similar findings were made by \cite{kaneko2022}. This could be because the current measures used to remove representation bias are superficial, as argued by \citet{gonen2019lipstick}.
\begin{table}[]
    \centering
    \renewcommand{\arraystretch}{1.10}
    \resizebox{0.9\textwidth}{!}{
    \begin{tabular}{lllrrr}
    \hline
         Attribute & Model & AUC & FPR\_gap & TPR\_gap & AUC\_gap  \\
    \hline
         \multirow{6}{*}{Gender} &
       ALBERT &
               \textbf{0.847} & {0.006} & {0.039}  & 0.004  \\

        & + upstream-sentDebias-gender  &
           {0.840} & {0.006} & {\color{teal} $\downarrow 0.032$} & 0.004   \\

      \cline{2-6}  &  BERT &
          {0.830} & {0.090} & {0.036} & 0.010  \\

         & + upstream-sentDebias-gender  &
          \textbf{0.841} & {\color{teal}$\downarrow0.011$} & {\color{red}$\uparrow0.049$} & \color{teal}$\downarrow0.006$ \\

       \cline{2-6}  
        & RoBERTa  &
          {0.851} & {0.005} & {0.032} & 0.011 \\

             & + upstream-sentDebias-gender  &
          \textbf{0.856} & {\color{red}$\uparrow0.006$} & {\color{teal}$\downarrow0.022$} & \color{teal}$\downarrow0.003$ \\
    \hline
         \multirow{6}{*}{Race} &
       ALBERT &
         \textbf{0.847} & {0.008} & {0.002} & 0.019  \\

        & + upstream-sentDebias-race  &
          {0.838} & {\color{teal} $\downarrow0.003$} & {\color{red} $\uparrow 0.003$} & \color{teal} $\downarrow0.013$ \\
          
      \cline{2-6}  &  BERT &
          \textbf{0.830} & {0.016} & {0.002} & 0.026  \\

        & + upstream-sentDebias-race  &
          {0.829} & {\color{red}$\uparrow0.021$} & {\color{red}$\uparrow0.005$} & \color{teal}$\downarrow0.024$ \\
          
       \cline{2-6}  
        & RoBERTa  &
          {0.851} & {0.003} & {0.011} & 0.021 \\

        & + upstream-sentDebias-race  &
          \textbf{0.854} & {\color{red}$\uparrow0.017$} & {\color{teal}$\downarrow0.009$} & 0.021 \\
         \hline 
        \multirow{6}{*}{Religion} &
       ALBERT &
          \textbf{0.847} & {0.010} & {0.109} & 0.020  \\

         & + upstream-sentDebias-religion  &
          {0.837} & {\color{red} $\uparrow0.019$} & {\color{teal}$\downarrow0.094$} & \color{teal}$\downarrow0.016$  \\

      \cline{2-6}  &  BERT &
          {0.830} & {0.008} & {0.063}  & 0.012\\

         & + upstream-sentDebias-religion  &
          \textbf {0.833} & {\color{red}$\uparrow.015$} & {\color{red}$\uparrow0.084$} & \color{red}$\uparrow0.017$ \\

       \cline{2-6}  
        & RoBERTa  &
          \textbf{0.851} & {0.022} & {0.160} & 0.027 \\

         & + upstream-sentDebias-religion  &
          {0.843} & {\color{teal}$\downarrow0.021$} & {\color{teal}$\downarrow0.100$} & \color{teal}$\downarrow0.003$  \\
          
         \hline 
  
    \end{tabular}
    } 
     \caption{Toxicity detection performance and fairness scores for all models before and after removing representation bias using SentDebias. \textbf{Bold} values refer to better AUC scores and better performance on toxicity detection. (\textcolor{red}{$\uparrow$}) denotes that the fairness metric score increased, and the fairness worsened. (\textcolor{teal}{$\downarrow$}) denotes that the fairness metric score decreased and the fairness improved. The word \textit{upstream} is used here to refer to removing bias from the models before fine-tuning them on the task of toxicity detection.}
    \label{tab:upstream_debias_fainress_jigsaw_dataset}
\end{table}
\subsection{Selection bias}
\label{sec:selection_bias}
\begin{figure}
    \centering
    \includegraphics[width=0.8\textwidth]{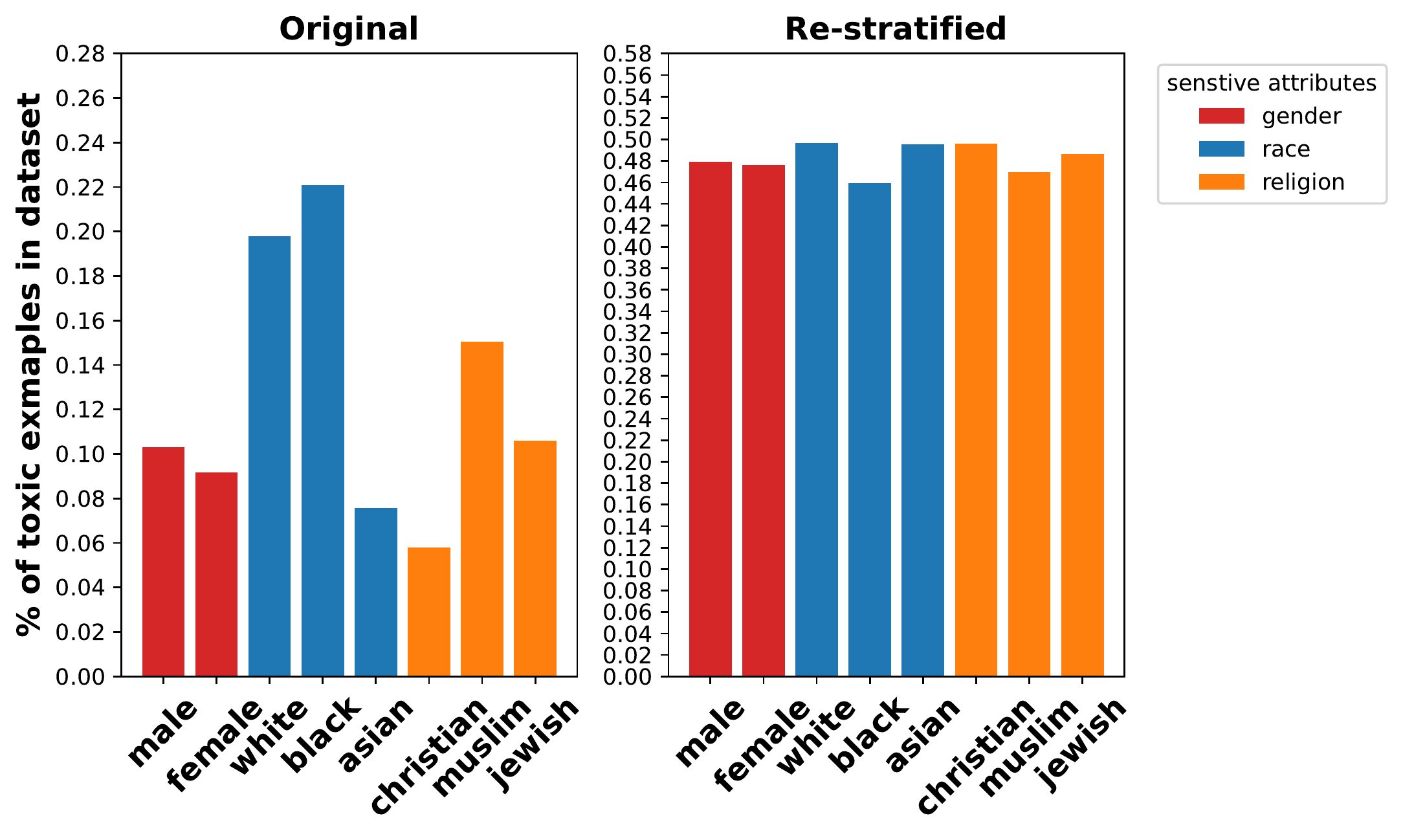}
    \caption{The percentage of positive (toxic) examples, for each identity group in the civil community training dataset in the original dataset (left) and after re-stratification (right).}
    \label{fig:selection_bias_positive_ratios}
\end{figure}

Selection bias \cite{hovy2021five}, also known as sample bias, is a result of non-representative observations in the training datasets used in downstream tasks \cite{shah-etal-2020-predictive}. For the task of toxicity detection, we interpreted selection bias as the over-representation of a certain identity group with the positive (toxic) label, as shown in \figurename~\ref{fig:selection_bias_positive_ratios} (Left). We measure selection bias in the civil community training dataset by measuring the difference in the ratios of the positive examples, between the marginalized and non-marginalized groups. Equation \ref{eq:selection-bias} shows how we to measure selection bias ($Selection_{g,\hat{g}}$) where ($N_{g}$) is the size of the data subset that is targeted at marginalized groups ($g$); ($N_{\hat{g}}$) is the size of the data subset that is targeted at non-marginalized groups ($\hat{g}$); ($N_{g, toxicity=1}$) is the number of toxic sentences that are targeted at marginalized groups; and ($N_{\hat{g}, toxicity=1}$) is the number of toxic sentences that are targeted at non-marginalized groups.  
The results indicate that the selection bias is the highest in the sensitive attribute of religion (0.077), followed by race (0.053), and finally gender (0.027). 

\begin{equation}\label{eq:selection-bias}
    Selection_{g, \hat{g}} = \left | \left (\frac{N_{g, toxicity=1}}{N_{g}} \right ) - \left (\frac{N_{\hat{g}, toxicity=1}}{N_{\hat{g}}} \right ) \right |
\end{equation}

To measure the impact of selection bias on the fairness of the toxicity detection task, we used the Pearson's correlation coefficient  ($\rho$) between the fairness scores measured by the different fairness metrics and selection bias scores in the civil community training dataset. We found that, for AlBERT-base, selection bias scores have a strong positive correlation with the fairness scores when measured as FPR\_gap ($\rho$ = 0.984), AUC\_gap ($\rho$ = 0.911), and  TPR\_gap ($\rho$ = 0.633). Similar correlation patterns were found in RoBERTa-base. As for BERT-base, there are weaker positive correlations with TPR\_gap and AUC\_gap and almost no correlation with FPR\_gap.

These results suggest that selection bias in the training datasets used for downstream tasks has a direct impact on the fairness of the examined language models, as evidenced by the positive correlations with the fairness of these models on the downstream task of toxicity detection, as measured by the different fairness metrics. 

\subsection{Selection bias removal}
\label{sec:selection-bias-removal}
According to \citet{shah-etal-2020-predictive}, to remove selection bias, a realignment in the sample distribution in the training dataset is required to minimize the mismatch in the class representation between the different identities. The authors in \citet{shah-etal-2020-predictive} list data re-stratification as a mitigation technique to remove selection bias and to match the ideal distribution of balanced class representations for the different identities. We followed this suggestion to have a balanced representation of positive and negative examples for the different marginalized and non-marginalized groups that we study in this section. 

Following the same methodology used in \citet{zmigrod-etal-2019-counterfactual}, data augmentation was used to create slightly altered examples to balance the class representations and add them to the civil community training dataset. Since the percentages of the positive examples for the different identity groups are small, ranging from 0.05 to 0.2, as shown in \figurename~\ref{fig:selection_bias_positive_ratios} (Left), we create more positive examples by slightly altering the existing positive examples, without changing the meaning, in the dataset using word substitutions. Word substitutions were generated using the NLPAUG tool\footnote{\url{https://github.com/makcedward/nlpaug}} that uses contextual word embeddings to find the word substitutions \cite{ma2019nlpaug}. After adding the synthesized positive examples to the civil community training dataset, the new re-stratified civil community training dataset contains 443,046 data items with balanced class representation, as shown in \figurename~\ref{fig:selection_bias_positive_ratios} (Right). The selection bias in the re-stratified civil community training dataset is reduced to 0.002, 0.019 and 0.017 for the gender, race, and religion sensitive attributes respectively. 

Then, the examined AlBERT, BERT, and RoBERTa models were fine-tuned on the new re-stratified civil community training dataset. Balancing the class representation in the dataset led to a reduction in the performance (AUC scores) of all three models (+ downstream-stratified-data), as shown in \tablename~\ref{tab:debias_selection_fairness_jigsaw_dataset}. This reduction in the AUC scores is a result of predicting more positive examples than the original models and in turn resulting in more false positive and more less true negative. 

\begin{table}
    \centering
    \renewcommand{\arraystretch}{1.10}
    \resizebox{0.9\textwidth}{!}{
    \begin{tabular}{lllrrr}
    \hline
         Attribute & Model & AUC & FPR\_gap & TPR\_gap & AUC\_gap  \\
    \hline
         \multirow{6}{*}{Gender} &
       ALBERT &
               \textbf{0.847} & {0.006} & {0.039}  & 0.004  \\

        & + downstream-stratified-data &
          {0.816} & {\color{teal} $\downarrow0.005$} & {\color{teal} $\downarrow0.003$} & \color{red}$\uparrow0.005$ \\

      \cline{2-6}  &  BERT &
          \textbf{0.830} & {0.090} & {0.036} & 0.010  \\
      
        & + downstream-stratified-data &
          {0.817} & {\color{teal}$\downarrow0.007$}  & {\color{teal}$\downarrow0.006$} & \color{teal}$\downarrow0.006$ \\

       \cline{2-6}  
        & RoBERTa  &
          \textbf{0.851} & {0.005} & {0.032} & 0.011 \\

        & + downstream-stratified-data &
          {0.842} & {\color{red}$\uparrow0.006$} & {\color{teal}$\downarrow0.005$} & \color{teal}$\downarrow0.002$ \\

     \hline
         \multirow{6}{*}{Race} &
       ALBERT &
         \textbf{0.847} & {0.008} & {0.002} & 0.019  \\

        & + downstream-stratified-data &
          {0.816} & {\color{red} $\uparrow0.022$} & {\color{red} $\uparrow0.026$} & \color{teal} $\downarrow0.008$ \\
         
      \cline{2-6}  &  BERT &
          \textbf{0.830} & {0.016} & {0.002} & 0.026  \\

        & + downstream-stratified-data &
          {0.817} & {\color{teal}$\downarrow0.010$} & {\color{red}$\uparrow0.018$} & \color{teal}$\downarrow0.008$  \\

     \cline{2-6}  
        & RoBERTa  &
          \textbf{0.851} & {0.003} & {0.011} & 0.021 \\

        & + downstream-stratified-data &
          {0.842} & {\color{red}$\uparrow.014$} & {0.011}  & \color{teal}$\downarrow0.014$ \\
          
         \hline 
        \multirow{6}{*}{Religion} &
       ALBERT &
          \textbf{0.847} & {0.010} & {0.109} & 0.020  \\

        & + downstream-stratified-data &
          {0.816} & {\color{red}$\uparrow0.030$} & {\color{teal}$\downarrow0.058$} & \color{teal}$\downarrow0$ \\

      \cline{2-6}  &  BERT &
          \textbf{0.830} & {0.008} & {0.063}  & 0.012\\

        & + downstream-stratified-data &
          {0.817} & {\color{red}$\uparrow0.020$} & {\color{teal}$\downarrow0.049$} & \color{teal}$\downarrow0.006$ \\

       \cline{2-6}  
        & RoBERTa  &
          \textbf{0.851} & {0.022} & {0.160} & 0.027 \\

        & + downstream-stratified-data &
          {0.842} & {\color{teal}$\downarrow0.019$} & {\color{teal}$\downarrow0.071$} & \color{teal}$\downarrow0.001$ \\

         \hline 
  
    \end{tabular}
    } 
    \caption{Toxicity detection performance and fairness scores for all models before and after removing selection bias. \textbf{Bold} values refer to higher AUC scores and better performance. (\textcolor{red}{$\uparrow$}) denotes that the fairness metric score increased and the fairness worsened. (\textcolor{teal}{$\downarrow$}) denotes that the fairness metric score decreased and the fairness improved. The word \textit{downstream} is used to explain that the bias removal technique is applied during fine-tuning the model on the downstream task of toxicity detection.}
    \label{tab:debias_selection_fairness_jigsaw_dataset}
\end{table}

To investigate the impact of selection bias removal on the fairness of the task of toxicity detection, we measure the fairness in all three examined models after fine-tuning them on the new re-stratified civil community dataset.
The fairness scores were analyzed for the examined sensitive attributes using the different fairness metrics for all the examined models. Results in \tablename~\ref{tab:debias_selection_fairness_jigsaw_dataset} showed that for the AUC\_gap metric, the fairness improved for all models and most of the sensitive attributes as evidenced by AlBERT (race, religion), BERT (gender, race, religion), and RoBERTa  (gender, race, religion). However, the results are inconsistent for the TPR\_gap or FPR\_gap across models or sensitive attributes. 
We speculate that TPR\_gap and FPR\_gap do not reflect the improvement in fairness as measured using the AUC\_gap metric because after removing selection bias, the models tend to predict more positives (false positive and true positive). The FPR and TPR increased for some identity groups (White and Black) and got reduced for other groups (Asian), which made the TPR\_gap and FPR\_gap increase. On the other hand, this does not happen with the AUC scores for the different identity groups, hence the AUC\_gap scores did not increase, which might be the case because the AUC scores and the AUC\_gap are threshold-agnostic metrics and don't directly rely on the models' false or true positive predictions.

When examining the cases where the fairness improved according to all the fairness metrics, we found only two cases out of nine. The first is when BERT is fine-tuned on the re-stratified training dataset, which led to improvement of the model's fairness regarding the gender sensitive attribute. The second is fine-tuning RoBERTa on the re-stratified training dataset, which led to improvement of the model's fairness regarding the religion sensitive attribute.


In summary, it was shown that selection bias is influential on the models' fairness on the task of toxicity detection and removing it by balancing the class representations for all the identity groups in the training dataset using data augmentation improved the models' fairness according to the AUC\_gap metric, but not according to all the examined fairness metrics.

\subsection{Overamplification bias}
\label{sec:overap_bias}
According to \citet{shah-etal-2020-predictive}, overamplification bias happens during LM training as LM models rely on small differences between sensitive attributes regarding an objective function and amplify these differences to be more pronounced in the predicted outcome. For the task of toxicity detection, overamplification bias could happen because certain identity groups exist more often within specific semantic contexts in the training datasets. For example, when an identity name, e.g., \say{Muslim} co-exists in the same sentence with the word \say{terrorism} more often than other identity names, e.g., \say{Buddist}. Even if the sentence does not contain any hate speech, e.g., ``\textit{Anyone could be a terrorist, not just muslims}'', the LM model will learn to pick this information up and amplify it, leading to the fine-tuned LM model predicting future sentences that contain the word "Muslim" as hateful. 

In \citet{zhao-etal-2017-men}, the authors propose a method to measure and mitigate overamplification bias when training models on biased corpora. The authors propose the RBA framework for reducing bias amplification in predictions. Their proposed method introduces corpus-level constraints so that gender indicators co-occur no more often together with elements of the prediction task than in the original training distribution \cite{zhao-etal-2017-men}. The aim of the proposed method is to limit the model bias to the bias in the dataset without amplification. This method would only be effective if the training dataset is not biased. So in this section, we aim to examine overamplification bias in the training dataset before it gets amplified during model training.

\begin{equation}
    Overamplification_{g, \hat{g}} = | N_{g} - N_{\hat{g}} |
    \label{eq:overampf_bias}
\end{equation}

\begin{figure}
    \centering
    \includegraphics[width=0.7\textwidth]{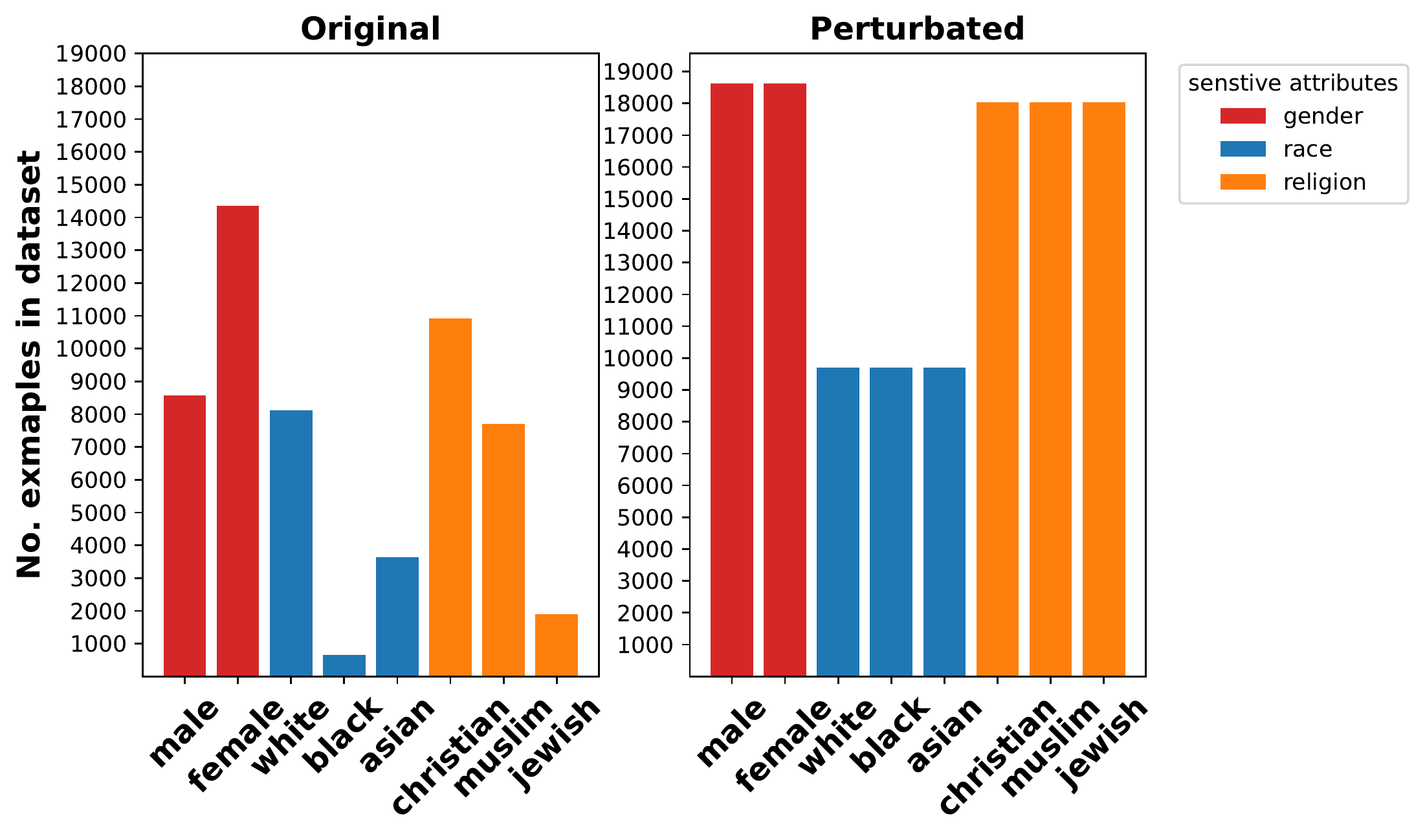}
    \caption{The number of examples, for each identity group in the civil community training dataset in the original dataset (left) and after perturbation (right).}
    \label{fig:size_biases}
\end{figure}

Equation~\ref{eq:overampf_bias} was used to measure overamplification bias in the civil community training dataset. To this end, we measured the differences between the number of examples targeted at marginalized vs. non-marginalized groups, as shown in \figurename~\ref{fig:size_biases} (Left).  Then the scores are normalized using the Max normalization \cite{LI2011256} where each value in $overamplification_{g, \hat{g}}$ for gender (5795), race(5795), and religion (6118.5) is divided by max value which is 6118.5. The reason behind using Max normalization is to avoid having a score of 0 which might be misleading in the context of bias. The different sizes mean that certain identity groups appear in more semantic contexts than others. These contexts could be positive or negative. 
The overamplification bias scores in the civil community training dataset for the sensitive attributes are 1.0 for religion, followed by 0.97 for race, and finally 0.94 for gender.

To investigate the impact of overamplification bias on the models' fairness on the downstream task of toxicity detection, we measured the Pearson correlation coefficient ($\rho$) between overamplification bias scores and the fairness scores measured using threshold-based and threshold-agnostic fairness metrics. A strong positive correlation was found for AlBERT-base between overamplification bias and fairness scores, as measured by FPR\_gap ($\rho$ = 0.988), AUC\_gap ($\rho$ = 0.921) and TPR\_gap ($\rho$ = 0.613). A similar pattern of correlations was found for RoBERTa-base. As for BERT-base, there are weaker positive correlations with TPR\_gap and AUC\_gap and almost no correlation with FPR\_gap.
These results suggest that overamplification bias in the civil community training dataset measured as the difference in the sentences that are targeted at the different groups might have a direct impact on the models' fairness, and that during fine-tuning these differences are amplified in a way that might then make a bigger impact on the models' fairness. Additionally, overamplification bias could also mean the amplification of selection bias introduced earlier in section~\ref{sec:selection_bias}.

\begin{figure}
    \centering
    \includegraphics[width=0.7\textwidth]{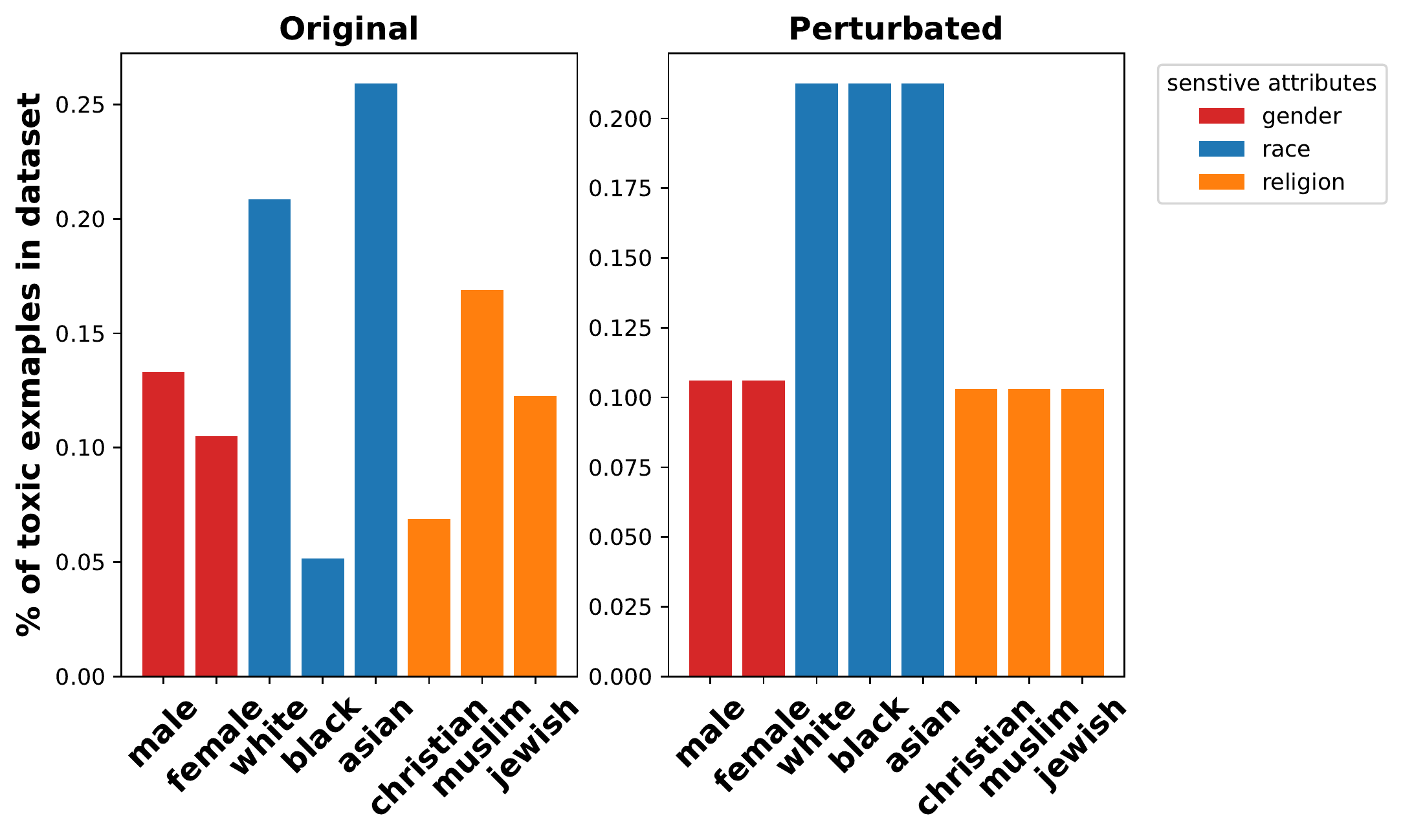}
    \caption{The percentage of positive (toxic) examples, for each identity group in the civil community
training dataset in the original dataset (left) and after perturbation (right).}
    \label{fig:percentage_size_biases}
\end{figure}

\subsection{Overamplification bias removal}
\label{sec:over-bias-removal}
To overcome overamplification bias, we followed the work of \citet{Webster-etal-2020-cda} where the authors propose to train the model on a training dataset with balanced semantic representations of the different identity groups using counterfactuals. To achieve that balance in the civil community training dataset, each identity, marginalized and non-marginalized, should be presented in similar semantic contexts so that the models would not associate certain semantic contexts with certain identity groups. 

To create the perturbations, the first attempt was to fine-tune a Text-to-Text model \cite{raffel2020exploring} on the PANDA dataset \cite{qian-etal-2022-perturbation} to automatically generate perturbations. We used the same values for the hyperparameters as shared in \citet{raffel2020exploring} and the Text-to-Text model achieved a ROUGE-2 score of 0.9, which is similar to the score reported in the original study (ROUGE-2=0.9)~\cite{qian-etal-2022-perturbation}. However, upon inspection of the perturbed text, it was found that the perturbed text is not consistently changing and that it does not perform well on religious or racial identities. Upon further inspection, we realized that the perturbed text in the PANDA dataset is inconsistent, and sometimes the perturbations are not correct. This is not picked up by the ROUGE-2 metric because it works by comparing the overlap of bi-grams between two sentences (original and perturbed) \cite{ng2015better}, which is not indicative of good performance in the task of perturbation generation in comparison to a task like text translation. Since the original and the perturbed sentences are similar except for words that describe identity groups, the ROUGE-2 metric gives high scores regardless of the quality of the generated perturbation.

To address this issue, we followed the same method used to create the balanced civil community fairness dataset, as explained in section~\ref{sec:extrinsic_bias}. To this end, perturbations (counterfactuals) were created for each sentence in the civil community training dataset. As discussed in section \ref{sec:balanced_toxiciy_fairness_dataset}, the most common nouns and adjectives in the civil community dataset suggest that there is no risk of \textit{Asymmetric Counterfactuals} which means we can create the perturbations for toxic and non-toxic sentences. So for the identity of Black people, in addition to the subset of sentences that are targeted at Black people, we created perturbations from the sentences that are targeted at White and Asian identities, replacing any references to White or Asian identity names with Black identity names. The same was done with the White identity. In addition to the sentences that are targeted at the White identity, perturbations were created from the sentences that are targeted at Black and Asian people, replacing the words that describe Asian or Black identities with identity words that describe the White identity. The same process was repeated for the Asian identity. In addition to the sentences that target Asian people, perturbations were created from the sentences that are targeted at Black and White people, replacing any reference to Black and White with identity words that describe Asian people. This way, it was ensured that all the different racial identities in the dataset are represented in the same way. The same process was repeated for the identity groups in the gender and religion sensitive attributes. The new distribution of identities is shown in \figurename~\ref{fig:size_biases} (Right). \figurename~\ref{fig:percentage_size_biases} shows how removing overamplification bias mitigated selection bias, as shown in \figurename~\ref{fig:percentage_size_biases} (Left), by balancing the ratios of positive examples for the different identity groups in the same sensitive attribute, as shown in \figurename~\ref{fig:percentage_size_biases} (Right).

The size of the balanced-perturbed civil community training dataset after generating the perturbations is 382,212 sentences and the ratio between the positive and the negative examples for each identity group within the same sensitive attribute is the same. For example, in the gender attribute, the ratio of the positive (toxic) examples in the male and female identity groups is 0.10, in the race attribute, the ratio of the positive examples for the Black, White and Asian groups is 0.2, and for the religion attribute, the ratio of the positive examples for the Muslim, Christian and Jewish groups is 0.10. It is worth noting that with similar ratios of positive examples between the different identity groups in the same sensitive attribute, selection bias in the new civil community training dataset is mitigated as well. Finally, we used the balanced perturbed civil community training dataset to fine-tune the examined models (+ downstream-perturbed-data). 

Since selection bias could also be amplified by the models during training, we perturbed the re-stratified civil community training dataset, as explained in section~\ref{sec:selection_bias}. This does not only guarantee balanced semantic contexts, but also a more balanced ratio between positive and negative examples in the civil community training dataset. The new perturbed-stratified dataset contains 841,814 sentences, where the ratio of the positive examples for the male and female identity groups is 0.48, the ratio of positive examples between Black, Asian, and White identity groups is 0.48, and the ratio of positive examples between the Muslim, Christian, and Jewish identity groups is 0.49.

As an alternative to using counterfactual and data perturbation to remove overamplification bias, we used SentDebias to remove the biased subspaces from the models as explained in section \ref{sec:sentDebias}. To this end , to remove overamplification bias, we remove the biased subspaces after fine-tuning the models on the civil community dataset (+ downstream-sentDebias). 

To investigate the impact of removing overamplification bias on the fairness of the toxicity detention task, we used the threshold-based and threshold-agnostic fairness metrics to measure the impact of the examined debiasing techniques for removing overamplification bias on the models' fairness. 
\begin{table}[]
    \centering
    \renewcommand{\arraystretch}{1.10}
    \resizebox{0.8\textwidth}{!}{
    \begin{tabular}{lllrrr}
    \hline
         Attribute & Model & AUC & FPR\_gap & TPR\_gap & AUC\_gap  \\
    \hline
         \multirow{12}{*}{Gender} &
       ALBERT &
               {0.847} & {0.006} & {0.039}  & 0.004  \\

        & + downstream-sentDebias-gender &
          {0.524} & {\color{teal}$\downarrow 0$}  & {\color{teal} $\downarrow 0.008$} & \color{red} $\uparrow 0.011$  \\

        & + downstream-perturbed-data &
          \textbf{0.848} & {\color{teal}$\downarrow 0.001$}  & {\color{teal} $\downarrow0.010$} & 0.004  \\

       &  + downstream-perturbed-stratified-data &
          {0.803} & {\color{teal} $\downarrow0.005$}  & {\color{teal} $\downarrow0.006$} & \color{red} $\uparrow0.008$ \\

      \cline{2-6}  &  BERT &
          {0.830} & {0.09} & {0.036} & 0.01  \\

        & + downstream-sentDebias-gender &
          {0.478} & {\color{teal}$\downarrow0$} & {\color{teal} $\downarrow0.001$} & \color{teal}$\downarrow0.004$  \\

        & + downstream-perturbed-data &
          \textbf{0.837} & {\color{teal}$\downarrow0.003$} & {\color{teal}$\downarrow0.005$} & \color{teal}$\downarrow0.003$  \\

       &  + downstream-perturbed-stratified-data &
          {0.810} & {\color{teal}$\downarrow0.003$} & {\color{teal}$\downarrow0.003$} & \color{teal}$\downarrow0.005$ \\

       \cline{2-6}  
        & RoBERTa  &
          {0.851} & {0.005} & {0.032} & 0.011 \\

        & + downstream-sentDebias-gender &
          {0.520} & {\color{red}$\uparrow0.015$} & {\color{teal}$\downarrow0.019$}  & \color{teal}$\downarrow0.004$ \\

        & + downstream-perturbed-data &
          \textbf{0.873} & {\color{teal}$\downarrow0.001$} & {\color{teal}$\downarrow0.009$} & \color{teal}$\downarrow0.002$ \\

       &  + downstream-perturbed-stratified-data &
          {0.825} & {\color{teal}$\downarrow0$}  & {\color{teal}$\downarrow0.005$} & \color{teal}$\downarrow0.007$ \\

    \hline
         \multirow{12}{*}{Race} &
       ALBERT &
         {0.847} & {0.008} & {0.002} & 0.019  \\

        & + downstream-sentDebias-race &
          {0.421} & {\color{teal} $\downarrow 0$} & {\color{red} $\uparrow0.004$} & \color{teal}$\downarrow0.001$  \\

        & + downstream-perturbed-data &
          \textbf{0.848} & {\color{teal} $\downarrow0.003$} & {\color{teal} $\downarrow0.001$} & \color{teal} $\downarrow0.003$ \\

       &  + downstream-perturbed-stratified-data &
          {0.803} & {\color{red}$\uparrow0.004$} & {0.002} & \color{teal} $\downarrow0.002$ \\

      \cline{2-6}  &  BERT &
          {0.830} & {0.016} & {0.002} & 0.026  \\

        & + downstream-sentDebias-race &
          {0.504} & {\color{teal}$\downarrow0$} & {\color{teal}$\downarrow0$} & \color{teal}$\downarrow0.002$ \\

        & + downstream-perturbed-data &
          \textbf{0.837} & {\color{teal}$\downarrow0.009$} & {\color{red}$\uparrow0.019$} & \color{teal}$\downarrow0.003$ \\

       &  + downstream-perturbed-stratified-data &
          {0.810} & {\color{teal}$\downarrow0.002$} & {0.002} & \color{teal}$\downarrow0.002$ \\

       \cline{2-6}  
        & RoBERTa  &
          {0.851} & {0.003} & {0.011} & 0.021 \\

        & + downstream-sentDebias-race &
          {0.561} & {\color{teal}$\downarrow0$} & {\color{teal}$\downarrow0$} & \color{teal}$\downarrow0.005$ \\

        & + downstream-perturbed-data &
          \textbf{0.873} & {\color{red}$\uparrow0.018$} & {\color{red}$\uparrow0.038$} & \color{teal}$\downarrow0.003$ \\

       &  + downstream-perturbed-stratified-data &
          {0.825} & {0.003} & {\color{teal}$\downarrow0.006$} & \color{teal}$\downarrow0.001$ \\

         \hline 
        \multirow{12}{*}{Religion} &
       ALBERT &
          {0.847} & {0.010} & {0.109} & 0.020  \\

        & + downstream-sentDebias-religion &
          {0.507} & {\color{teal}$\downarrow0.004$} & {\color{teal} $\downarrow0$}  & \color{teal} $\downarrow0.002$  \\

        & + downstream-perturbed-data &
          \textbf{0.848} & {\color{teal} $\downarrow0.002$} & {\color{teal}$\downarrow0.011$} & \color{teal}$\downarrow0.001$ \\

       &  + downstream-perturbed-stratified-data &
          {0.803} & {\color{teal}$\downarrow0$} & {\color{teal}$\downarrow 0.002$} & \color{teal} $\downarrow0.002$  \\

      \cline{2-6}  &  BERT &
          {0.830} & {0.008} & {0.063}  & 0.012\\

        & + downstream-sentDebias-religion &
          {0.447} & {\color{teal}$\downarrow0$} & {\color{teal}$\downarrow0$} & \color{red}$\uparrow0.030$ \\

        & + downstream-perturbed-data &
          \textbf{0.837} & {\color{teal}$\downarrow0.002$} & {\color{teal}$\downarrow0.011$} & \color{teal}$\downarrow0.001$ \\

       &  + downstream-perturbed-stratified-data &
          {0.810} & {\color{teal}$\downarrow0$} & {\color{teal}$\downarrow0.001$} & \color{teal}$\downarrow0.003$ \\

       \cline{2-6}  
        & RoBERTa  &
          {0.851} & {0.022} & {0.160} & 0.027 \\

        & + downstream-sentDebias-religion &
          {0.523} & {\color{teal}$\downarrow0$}  & {\color{teal}$\downarrow0$}  & \color{teal}$\downarrow0$ \\

        & + downstream-perturbed-data &
          \textbf{0.873} & {\color{teal}$\downarrow0.001$} & {\color{teal}$\downarrow0.003$} & \color{teal}$\downarrow0.002$ \\

       &  + downstream-perturbed-stratified-data &
          {0.825} & {\color{teal}$\downarrow0.001$} & {\color{teal}$\downarrow0.004$} & \color{teal}$\downarrow0.001$ \\
         \hline 
  
    \end{tabular}
    } 
    \caption{Toxicity detection performance and fairness scores for all models before and after overamplification bias removal. \textbf{Bold} values refer to higher AUC scores and better performance . (\textcolor{red}{$\uparrow$}) denotes that the fairness score increased and the fairness worsened. (\textcolor{teal}{$\downarrow$}) denotes that the fairness score decreased and the fairness improved.The word \textit{downstream} is used to explain that the bias removal technique is applied during fine-tuning the model on the downstream task of toxicity detection.}
    \label{tab:debias_overampflication_performance_jigsaw_dataset}
\end{table}
\begin{table}[]
    \centering
    \renewcommand{\arraystretch}{1.05}
    \resizebox{0.85\textwidth}{!}{
    \begin{tabular}{lllrrr}
    \hline
         Attribute & Model & AUC & FPR\_gap & TPR\_gap & AUC\_gap  \\
    \hline
         \multirow{21}{*}{Gender} &
       ALBERT &
               {0.847} & {0.006} & {0.039}  & 0.004  \\

        & + upstream-sentDebias-gender  &
           {0.840} & {0.006} & {\color{teal} $\downarrow 0.032$} & 0.004   \\

        & + downstream-sentDebias-gender &
          {0.524} & {\color{teal}$\downarrow 0$}  & {\color{teal} $\downarrow 0.008$} & \color{red} $\uparrow 0.011$  \\

        & + downstream-perturbed-data &
          \textbf{0.848} & {\color{teal}$\downarrow 0.001$}  & {\color{teal} $\downarrow0.01$} & 0.004  \\

        & + downstream-stratified-data &
          {0.816} & {\color{teal} $\downarrow0.005$} & {\color{teal} $\downarrow0.003$} & \color{red}$\uparrow0.005$ \\

       &  + downstream-perturbed-stratified-data &
          {0.803} & {\color{teal} $\downarrow0.005$}  & {\color{teal} $\downarrow0.006$} & \color{red} $\uparrow0.008$ \\

        & + upstream-sentDebias-gender- downstream-all-data-debias&
           {0.792} & {\color{teal}$\downarrow0.001$}  & {\color{teal} $\downarrow0.005$} & \color{red} $\uparrow0.008$  \\

      \cline{2-6}  &  BERT &
          {0.83} & {0.09} & {0.036} & 0.01  \\

         & + upstream-sentDebias-gender  &
          \textbf{0.841} & {\color{teal}$\downarrow0.011$} & {\color{red}$\uparrow0.049$} & \color{teal}$\downarrow0.006$ \\

        & + downstream-sentDebias-gender &
          {0.478} & {\color{teal}$\downarrow0$} & {\color{teal} $\downarrow0.001$} & \color{teal}$\downarrow0.004$  \\

        & + downstream-perturbed-data &
          {0.837} & {\color{teal}$\downarrow0.003$} & {\color{teal}$\downarrow0.005$} & \color{teal}$\downarrow0.003$  \\

        & + downstream-stratified-data &
          {0.817} & {\color{teal}$\downarrow0.007$}  & {\color{teal}$\downarrow0.006$} & \color{teal}$\downarrow0.006$ \\

       &  + downstream-perturbed-stratified-data &
          {0.810} & {\color{teal}$\downarrow0.003$} & {\color{teal}$\downarrow0.003$} & \color{teal}$\downarrow0.005$ \\

        & + upstream-sentDebias-gender- downstream-all-data-debias&
          {0.791} & {\color{teal}$\downarrow0$} & {\color{teal}$\downarrow0.003$} & \color{teal}$\downarrow0.002$ \\
       \cline{2-6}  
        & RoBERTa  &
          {0.851} & {0.005} & {0.032} & 0.011 \\

             & + upstream-sentDebias-gender  &
          {0.856} & {\color{red}$\uparrow0.006$} & {\color{teal}$\downarrow0.022$} & \color{teal}$\downarrow0.003$ \\

        & + downstream-sentDebias-gender &
          {0.520} & {\color{red}$\uparrow0.015$} & {\color{teal}$\downarrow0.019$}  & \color{teal}$\downarrow0.004$ \\

        & + downstream-perturbed-data &
          \textbf{0.873} & {\color{teal}$\downarrow0.001$} & {\color{teal}$\downarrow0.009$} & \color{teal}$\downarrow0.002$ \\

        & + downstream-stratified-data &
          {0.842} & {\color{red}$\uparrow0.006$} & {\color{teal}$\downarrow0.005$} & \color{teal}$\downarrow0.002$ \\

       &  + downstream-perturbed-stratified-data &
          {0.825} & {\color{teal}$\downarrow0$}  & {\color{teal}$\downarrow0.005$} & \color{teal}$\downarrow0.007$ \\

        & + upstream-sentDebias-gender-downstream-all-data-debias &
          {0.837} & {\color{teal}$\downarrow0.001$} & {\color{teal}$\downarrow0.003$} & \color{teal}$\downarrow0.004$ \\
    \hline
         \multirow{21}{*}{Race} &
       ALBERT &
         {0.847} & {0.008} & {0.002} & 0.019  \\

        & + upstream-sentDebias-race  &
          {0.838} & {\color{teal} $\downarrow0.003$} & {\color{red} $\uparrow 0.003$} & \color{teal} $\downarrow0.013$ \\

        & + downstream-sentDebias-race &
          {0.421} & {\color{teal} $\downarrow 0$} & {\color{red} $\uparrow0.004$} & \color{teal}$\downarrow0.001$  \\

        & + downstream-perturbed-data &
          \textbf{0.848} & {\color{teal} $\downarrow0.003$} & {\color{teal} $\downarrow0.001$} & \color{teal} $\downarrow0.003$ \\

        & + downstream-stratified-data &
          {0.816} & {\color{red} $\uparrow0.022$} & {\color{red} $\uparrow0.026$} & \color{teal} $\downarrow0.008$ \\

       &  + downstream-perturbed-stratified-data &
          {0.803} & {\color{red}$\uparrow0.004$} & {0.002} & \color{teal} $\downarrow0.002$ \\

        & + upstream-sentDebias-race-downstream-all-data-debias&
          {0.817} & {\color{teal}$\downarrow0.001$} & {\color{red} $\uparrow0.017$} & \color{teal} $\downarrow0.001$ \\
          
      \cline{2-6}  &  BERT &
          {0.830} & {0.016} & {0.002} & 0.026  \\

        & + upstream-sentDebias-race  &
          {0.829} & {\color{red}$\uparrow0.021$} & {\color{red}$\uparrow0.005$} & \color{teal}$\downarrow0.024$ \\

        & + downstream-sentDebias-race &
          {0.504} & {\color{teal}$\downarrow0$} & {\color{teal}$\downarrow0$} & \color{teal}$\downarrow0.002$ \\

        & + downstream-perturbed-data &
          \textbf{0.837} & {\color{teal}$\downarrow0.009$} & {\color{red}$\uparrow0.019$} & \color{teal}$\downarrow0.003$ \\

        & + downstream-stratified-data &
          {0.817} & {\color{teal}$\downarrow0.010$} & {\color{red}$\uparrow0.018$} & \color{teal}$\downarrow0.008$  \\

       &  + downstream-perturbed-stratified-data &
          {0.810} & {\color{teal}$\downarrow0.002$} & {0.002} & \color{teal}$\downarrow0.002$ \\

        & + upstream-sentDebias-race-downstream-all-data-debias &
          {0.815} & {\color{teal}$\downarrow0.005$} & {0.002} & \color{teal}$\downarrow0$ \\
       \cline{2-6}  
        & RoBERTa  &
          {0.851} & {0.003} & {0.011} & 0.021 \\

        & + upstream-sentDebias-race  &
          {0.854} & {\color{red}$\uparrow0.017$} & {\color{teal}$\downarrow0.009$} & 0.021 \\

        & + downstream-sentDebias-race &
          {0.561} & {\color{teal}$\downarrow0$} & {\color{teal}$\downarrow0$} & \color{teal}$\downarrow0.005$ \\

        & + downstream-perturbed-data &
          \textbf{0.873} & {\color{red}$\uparrow0.018$} & {\color{red}$\uparrow0.038$} & \color{teal}$\downarrow0.003$ \\

        & + downstream-stratified-data &
          {0.842} & {\color{red}$\uparrow.014$} & {0.011}  & \color{teal}$\downarrow0.014$ \\

       &  + downstream-perturbed-stratified-data &
          {0.825} & {0.003} & {\color{teal}$\downarrow0.006$} & \color{teal}$\downarrow0.001$ \\

        & + upstream-sentDebias-race-downstream-all-data-debias&
          {0.842} & {\color{red}$\uparrow0.006$} & {\color{red}$\uparrow0.014$} & \color{teal}$\downarrow0.003$ \\
         \hline 
        \multirow{21}{*}{Religion} &
       ALBERT &
          {0.847} & {0.010} & {0.109} & 0.020  \\

         & + upstream-sentDebias-religion  &
          {0.837} & {\color{red} $\uparrow0.019$} & {\color{teal}$\downarrow0.094$} & \color{teal}$\downarrow0.016$  \\

        & + downstream-sentDebias-religion &
          {0.507} & {\color{teal}$\downarrow0.004$} & {\color{teal} $\downarrow0$}  & \color{teal} $\downarrow0.002$  \\

        & + downstream-perturbed-data &
          \textbf{0.848} & {\color{teal} $\downarrow0.002$} & {\color{teal}$\downarrow0.011$} & \color{teal}$\downarrow0.001$ \\

        & + downstream-stratified-data &
          {0.816} & {\color{red}$\uparrow0.03$} & {\color{teal}$\downarrow0.058$} & \color{teal}$\downarrow0$ \\

       &  + downstream-perturbed-stratified-data &
          {0.803} & {\color{teal}$\downarrow0$} & {\color{teal}$\downarrow 0.002$} & \color{teal} $\downarrow0.002$  \\

        & + upstream-sentDebias-religion-downstream-all-data-debias&
          {0.811}  & {\color{teal}$\downarrow0.001$} & {\color{teal} $\downarrow0.013$} & \color{teal}$\downarrow0.003$  \\

      \cline{2-6}  &  BERT &
          {0.830} & {0.008} & {0.063}  & 0.012\\

         & + upstream-sentDebias-religion  &
          {0.833} & {\color{red}$\uparrow.015$} & {\color{red}$\uparrow0.084$} & \color{red}$\uparrow0.017$ \\

        & + downstream-sentDebias-religion &
          {0.447} & {\color{teal}$\downarrow0$} & {\color{teal}$\downarrow0$} & \color{red}$\uparrow0.03$ \\

        & + downstream-perturbed-data &
          \textbf{0.837} & {\color{teal}$\downarrow0.002$} & {\color{teal}$\downarrow0.011$} & \color{teal}$\downarrow0.001$ \\

        & + downstream-stratified-data &
          {0.817} & {\color{red}$\uparrow0.02$} & {\color{teal}$\downarrow0.049$} & \color{teal}$\downarrow0.006$ \\

       &  + downstream-perturbed-stratified-data &
          {0.810} & {\color{teal}$\downarrow0$} & {\color{teal}$\downarrow0.001$} & \color{teal}$\downarrow0.003$ \\

        & + upstream-sentDebias-religion-downstream-all-data-debias&
          {0.820} & {\color{teal}$\downarrow0$} & {\color{teal}$\downarrow0.005$} & \color{teal}$\downarrow0.003$ \\
       \cline{2-6}  
        & RoBERTa  &
          {0.851} & {0.022} & {0.16} & 0.027 \\

         & + upstream-sentDebias-religion  &
          {0.843} & {\color{teal}$\downarrow0.021$} & {\color{teal}$\downarrow0.10$} & \color{teal}$\downarrow0.003$  \\

        & + downstream-sentDebias-religion &
          {0.523} & {\color{teal}$\downarrow0$}  & {\color{teal}$\downarrow0$}  & \color{teal}$\downarrow0$ \\

        & + downstream-perturbed-data &
          \textbf{0.873} & {\color{teal}$\downarrow0.001$} & {\color{teal}$\downarrow0.003$} & \color{teal}$\downarrow0.002$ \\

        & + downstream-stratified-data &
          {0.842} & {\color{teal}$\downarrow0.019$} & {\color{teal}$\downarrow0.071$} & \color{teal}$\downarrow0.001$ \\

       &  + downstream-perturbed-stratified-data &
          {0.825} & {\color{teal}$\downarrow0.001$} & {\color{teal}$\downarrow0.004$} & \color{teal}$\downarrow0.001$ \\

        & + upstream-sentDebias-religion-downstream-all-data-debias &
          {0.834} & {\color{teal}$\downarrow0$} & {\color{teal}$\downarrow0.006$} & \color{teal}$\downarrow0.007$ \\
         \hline 
  
    \end{tabular}
    } 
    \caption{Summary of the performance and fairness scores for all models before and after applying different bias removal methods to remove different sources of bias. \textbf{Bold} values refer to higher AUC scores. (\textcolor{red}{$\uparrow$}) denotes that the fairness score increased and the fairness worsened. (\textcolor{teal}{$\downarrow$}) denotes that the fairness score decreased and the fairness improved. The word \textit{upstream} is used here to refer to removing bias from the models before fine-tuning them on the task of toxicity detection. The word \textit{downstream} is used to explain that the bias removal technique is applied during fine-tuning the model on the downstream task of toxicity detection.}
    \label{tab:debias_performance_jigsaw_dataset}
\end{table}
\begin{itemize}
\setlength\itemsep{1em}
    \item \textbf{Downstream-SentDebias:} Starting with the impact of removing the biased subspaces from the fine-tuned models, it is shown that the performance of the models after removing the biased representations is much worse, almost random, with the AUC scores close to 0.5 as shown in \Cref{tab:debias_overampflication_performance_jigsaw_dataset} (+ downstream-sentDebias). This is expected since the models lost a lot of information related to toxicity along with the biased subspaces. To simplify the result's analysis, we examined the impact of removing a certain type of bias on the fairness of the corresponding sensitive attribute, as explained in section~\ref{sec:intrinsic_bias}. 
    
    The results show that removing the biased subspaces after fine-tuning the models led to improved fairness in all the models according to all fairness metrics for almost all the sensitive attributes. However, these results are misleading. After analyzing the prediction probabilities of the models after removing the biased subspaces, we found that in most of the models, the number of positive predictions is either immensely reduced or immensely increased. This results in very low false positive rates and very low true positive rates, or very high false positive rates and very high true positive rates, resulting in minimal TPR\_gap and FPR\_gap scores ($\approx 0$). These results suggest that removing the biased subspace from fine-tuned models to mitigate overamplification bias falsely improves the models' fairness while it deteriorates their performance.
    
    \item \textbf{Data perturbation:} As for the impact of fine-tuning the models on perturbed datasets, the results in \Cref{tab:debias_overampflication_performance_jigsaw_dataset} (+ downstream-perturbed-data) show that the performance slightly improved for all the models. Fine-tuning the models on the perturbed data made the models predict more positives, true positives and false positives, without hurting the true negatives much, which is the case with the other examined debiasing methods. 
    
    An examination of the fairness scores after fine-tuning the models of the perturbed dataset shows that the different fairness metrics agree, in almost all the models for most of the sensitive attributes, that fine-tuning the models on the perturbed dataset improves the fairness. These results also suggest that mitigating overamplification bias by fine-tuning language models on perturbed datasets with balanced semantic contexts for different identity groups improved the fairness and the performance of the examined language models.
    
    \item \textbf{Perturbed-re-stratified data:} When the models were fine-tuned on the perturbed-re-stratified civil community dataset, the performance was slightly worse for all three models, as shown in \tablename~\ref{tab:debias_overampflication_performance_jigsaw_dataset} (+ downstream-perturbed-stratified-data). Like most of the other debiasing techniques, fine-tuning the perturbed re-stratified data caused the model to predict more positives, but especially more false positives in AlBERT and RoBERTa.  
    
    An examination of the fairness scores shows that the AUC\_gap consistently improved across all models and for almost all the sensitive attributes, AlBERT (race, religion),  BERT (gender, race, religion), and RoBERTa  (gender, race, religion). The results for FPR\_gap and TPR\_gap are not as consistent, but still improved for most of the sensitive attributes and models. These improvements are better than removing only selection bias by fine-tuning the models on re-stratified data. Yet not as good as fine-tuning the models on only perturbed-data, which improved the models' fairness more consistently.
\end{itemize}

\subsection{Multibiases}
\label{sec:multibiases}
Since all the examined sources of bias do not happen isolated from one another, we examined the impact of implementing all the proposed methods to remove bias from the different sources, as explained in sections \ref{sec:sentDebias}, \ref{sec:selection-bias-removal}, and \ref{sec:over-bias-removal}. To this end, we fine-tuned the models after removing the representation bias (upstream) using SentDebias on the perturbed-re-stratified civil community training dataset to remove both selection and overamplification biases (downstream). The AUC scores were slightly worse compared to the original models, as shown in \tablename~\ref{tab:debias_performance_jigsaw_dataset} (+upstream-sentDebias-gender-downstream-all-data-debias). 

The results indicate that
, similar to previous results, removing different sources of bias made the model predict more positives (false positives and true positives). Fore exmaple, removing gender bias in ALBERT (+ upstream-sentDebias-gender-downstream-all-data-debias), BERT (+ upstream-sentDebias-gender-downstream-all-data-debias), and RoBERTa (+ upstream-sentDebias-gender-downstream-all-data-debias) led to worse AUC scores because the number of false negatives also increased, which is similar to removing racial and religious bias. 
The fairness scores show improvement across most of the inspected models, fairness metrics, and sensitive attributes. The results show that removing different sources of bias, in most cases, improved the fairness scores ways similar to the results of fine-tuning the models on perturbed-re-stratified datasets. For example, the pattern of fairness improvement scores is the same in Gender (AlBERT, BERT, RoBERTa), Race (BERT), Religion (ALBET, BERT, RoBERTa). 

On the other hand, there are barely any similarities to the fairness improvement pattern between removing all sources of bias and removing representation bias. These results suggest that removing the downstream sources of bias (selection and overamplification) has a stronger impact on the models' fairness than the upstream sources of bias (representation bias). A similar finding was made by \citet{steed-etal-2022-upstream}.

\section{Discussion}
After investigating the impact of the different sources of bias and their removal on the downstream task of toxicity detection, we further analyse and summarize our results to answer our research questions.

\subsection{How impactful are the different sources of bias on the fairness of language models on the downstream task of toxicity detection?}
\label{sec:RQ1}
The results of the previous sections show that there is a positive correlation between representation bias, selection bias, and overamplification bias and the fairness scores measured by the different fairness metrics. This suggests that all the inspected sources of bias have an impact on the fairness of the inspected language models on the downstream task of toxicity detection. To find out the most impactful source of bias, we compared the strength of the correlation between the different sources of bias and the models' fairness. The CrowS-Pairs scores were used to measure representation bias, as CrowS-Pairs is the only metric that correlates positively with all the different fairness metrics, as explained in section~\ref{sec:intrinsic_bias}. For selection and overamplification sources of bias, we used the scores reported in sections~\ref{sec:selection_bias} and \ref{sec:overap_bias} respectively.

\begin{table}
\centering
    \renewcommand{\arraystretch}{1.10}
    \resizebox{0.5\textwidth}{!}{
\begin{tabular}{llll}
\hline
\multicolumn{4}{c}{\textbf{AlBERT}}                                                                                                                                   \\ \hline
\multicolumn{1}{l|}{}                                                                       & \multicolumn{3}{c}{Fairness}                                                  \\ \hline
\multicolumn{1}{l|}{Source of bias}                                                         & \multicolumn{1}{l|}{FPR\_gap} & \multicolumn{1}{l|}{TPR\_gap}       & AUC\_gap \\ \hline
\multicolumn{1}{l|}{\begin{tabular}[c]{@{}l@{}}Representation\\ (crowS-Pairs)\end{tabular}} & \multicolumn{1}{l|}{0.466}    & \multicolumn{1}{l|}{\textbf{0.999}}          & 0.233    \\ \hline
\multicolumn{1}{l|}{Selection}                                                              & \multicolumn{1}{l|}{0.984}    & \multicolumn{1}{l|}{0.633}         & 0.911    \\ \hline
\multicolumn{1}{l|}{Overampflication}                                                       & \multicolumn{1}{l|}{\textbf{0.988}}    & \multicolumn{1}{l|}{{0.613}}          & \textbf{0.921}    \\ \hline
\multicolumn{4}{c}{\textbf{BERT}}                                                                                                                                          \\ \hline
\multicolumn{1}{l|}{}                                                                       & \multicolumn{3}{c}{Fairness}                                                  \\ \hline
\multicolumn{1}{l|}{Source of bias}                                                         & \multicolumn{1}{l|}{FPR\_gap} & \multicolumn{1}{l|}{TPR\_gap}       & AUC\_gap \\ \hline
\multicolumn{1}{l|}{\begin{tabular}[c]{@{}l@{}}Representation\\ (crowS-Pairs)\end{tabular}} & \multicolumn{1}{l|}{-0.536}   & \multicolumn{1}{l|}{\textbf{0.819}} & -0.369   \\ \hline
\multicolumn{1}{l|}{Selection}                                                              & \multicolumn{1}{l|}{{-0.037}}    & \multicolumn{1}{l|}{0.418}         & {0.150}    \\ \hline
\multicolumn{1}{l|}{Overampflication}                                                       & \multicolumn{1}{l|}{\textbf{-0.011}}    & \multicolumn{1}{l|}{0.395}          & \textbf{0.175}    \\ \hline
\multicolumn{4}{c}{\textbf{RoBERTa}}                                                                                                                                   \\ \hline
\multicolumn{1}{l|}{}                                                                       & \multicolumn{3}{c}{Fairness}                                                  \\ \hline
\multicolumn{1}{l|}{Source of bias}                                                         & \multicolumn{1}{l|}{FPR\_gap} & \multicolumn{1}{l|}{TPR\_gap}       & AUC\_gap \\ \hline
\multicolumn{1}{l|}{\begin{tabular}[c]{@{}l@{}}Representation\\ (crowS-Pairs)\end{tabular}} & \multicolumn{1}{l|}{\textbf{0.972}}    & \multicolumn{1}{l|}{\textbf{0.980}}          & 0.555    \\ \hline
\multicolumn{1}{l|}{Selection}                                                              & \multicolumn{1}{l|}{0.809}    & \multicolumn{1}{l|}{0.785}          & 0.992    \\ \hline
\multicolumn{1}{l|}{Overampflication}                                                       & \multicolumn{1}{l|}{{0.794}}    & \multicolumn{1}{l|}{{0.770}}          & \textbf{0.995}    \\ \hline
\end{tabular}}
\caption{The Pearson correlation coefficient ($\rho$) between different sources of bias and fairness metrics in all the models. \textbf{Bold} values refer to the highest ($\rho$) values and hence the strongest correlation.}
\label{tab:source_bias_vs_fairness}
\end{table}

The results shown in \tablename~\ref{tab:source_bias_vs_fairness} indicate that correlation coefficients are high for both selection and overamplification bias in both AlBERT and BERT models. As for the RoBERTa model, representation bias has the highest correlation, which could be the case because RoBERTa is the most representation-biased model for all the sensitive attributes (gender, race, religion) when measured using the Crows-pairs metric, as shown in \tablename~\ref{tab:intrinsic_bias_scores}.
To summarize these findings and answer the research question, the results suggest that downstream bias sources of bias (selection bias and overamplification bias) is the most influential bias in comparison to upstream bias (representation bias). The results also suggest that overamplification bias is more impactful, as evidenced by the higher correlation coefficients, than representation and selection sources of bias. To have more conclusive answers, we investigated the most effective debiasing method as an indicator of the most impactful source of bias.

\subsection{What is the impact of removing the different sources of bias on the fairness of the downstream task of toxicity detection?}
\label{sec:RQ2}
To answer this research question, we summarized the findings on the impact of removing the different sources of bias on the models' fairness, as discussed in sections \ref{sec:sentDebias}, \ref{sec:selection-bias-removal}, \ref{sec:over-bias-removal}, and \ref{sec:multibiases}. 
The results in \tablename~\ref{tab:debiasisng-tech-summary} show that the most effective debiasing method that improved fairness according to all the used fairness metrics in most of the models and sensitive attributes is removing overamplification bias. These results support the finding from the previous research question that the most impactful source of bias is overamplification and removing it is the most effective on the models' fairness. The results also show that fine-tuning language models on perturbed data with balanced contextual semantic representation is more effective than training on perturbed re-stratified data. Furthermore, fine-tuning the models on perturbed data also addresses selection bias, as the ratio of positive examples is the same for all identity groups in the same sensitive attributes. It is also important to mention that removing downstream sources of bias (selection bias and overamplification bias) is more effective than removing upstream representation bias. This finding was also made by \citet{kaneko2022, steed-etal-2022-upstream}.
\begin{table}
    \renewcommand{\arraystretch}{1.2}
    \resizebox{1\textwidth}{!}{
\begin{tabular}{l|ccc|ccc|ccc}
\hline
                                                           & \multicolumn{3}{c|}{AlBERT-base}                                    & \multicolumn{3}{c|}{BERT-base}                                      & \multicolumn{3}{c}{RoBERTa-base}                                   \\ \hline
Debias approach & \multicolumn{1}{l|}{gender} & \multicolumn{1}{l|}{race}  & religion & \multicolumn{1}{l|}{gender} & \multicolumn{1}{l|}{race}  & religion & \multicolumn{1}{l|}{gender} & \multicolumn{1}{l|}{race}  & religion \\ \hline
Upstream-SentDebias                                        & \multicolumn{1}{l|}{\color{red} \xmark}  & \multicolumn{1}{l|}{\color{red} \xmark} & \color{red} \xmark    & \multicolumn{1}{l|}{\color{red} \xmark}  & \multicolumn{1}{l|}{\color{red} \xmark} & \color{red} \xmark    & \multicolumn{1}{l|}{\color{red} \xmark}  & \multicolumn{1}{l|}{\color{red} \xmark} & \color{teal} \cmark    \\ \hline
Downstream-SentDebias                                      & \multicolumn{1}{l|}{\color{red} \xmark}  & \multicolumn{1}{l|}{\color{red} \xmark} & \color{teal} \cmark    & \multicolumn{1}{l|}{\color{teal} \cmark}  & \multicolumn{1}{l|}{\color{teal} \cmark} & \color{red} \xmark    & \multicolumn{1}{l|}{\color{red} \xmark}  & \multicolumn{1}{l|}{\color{teal} \cmark} & \color{teal} \cmark    \\ \hline
Downstream-perturbed-data                                  & \multicolumn{1}{l|}{\color{red} \xmark}  & \multicolumn{1}{l|}{\color{teal} \cmark} & \color{teal} \cmark    & \multicolumn{1}{l|}{\color{teal} \cmark}  & \multicolumn{1}{l|}{\color{red} \xmark} & \color{teal} \cmark    & \multicolumn{1}{l|}{\color{teal} \cmark}  & \multicolumn{1}{l|}{\color{red} \xmark} & \color{teal} \cmark    \\ \hline
Downstream-stratified-data                                 & \multicolumn{1}{l|}{\color{red} \xmark}  & \multicolumn{1}{l|}{\color{red} \xmark} & \color{red} \xmark    & \multicolumn{1}{l|}{\color{teal} \cmark}  & \multicolumn{1}{l|}{\color{red} \xmark} & \color{red} \xmark    & \multicolumn{1}{l|}{\color{red} \xmark}  & \multicolumn{1}{l|}{\color{red} \xmark} & \color{teal} \cmark    \\ \hline
Downstream-perturbed-stratified-data                       & \multicolumn{1}{l|}{\color{red} \xmark}  & \multicolumn{1}{l|}{\color{red} \xmark} & \color{teal} \cmark    & \multicolumn{1}{l|}{\color{teal} \cmark}  & \multicolumn{1}{l|}{\color{red} \xmark} & \color{teal} \cmark    & \multicolumn{1}{l|}{\color{teal} \cmark}  & \multicolumn{1}{l|}{\color{red} \xmark} & \color{teal} \cmark    \\ \hline
Upstream-sentDebias-Downstream-all-data-debias             & \multicolumn{1}{l|}{\color{red} \xmark}  & \multicolumn{1}{l|}{\color{red} \xmark} & \color{teal} \cmark    & \multicolumn{1}{l|}{\color{teal} \cmark}  & \multicolumn{1}{l|}{\color{red} \xmark} & \color{teal} \cmark    & \multicolumn{1}{l|}{\color{teal} \cmark}  & \multicolumn{1}{l|}{\color{red} \xmark} & \color{teal} \cmark    \\ \hline
\end{tabular}}
\caption{Summary of the most effective debiasing method according to all the fairness metrics for all the models and all the sensitive attributes.}
\label{tab:debiasisng-tech-summary}
\end{table}
Removing the biased subspaces after fine-tuning (Downstream-SentDebias) is effective in some cases, like AlBERT (religion), BERT (gender, race), RoBERTa  (race, religion). However, using this technique leads to poor performance. Consequently, we do not recommend using this debiasing technique, as it is important to find the right trade-off between performance and fairness.
Removing selection bias by fine-tuning the models on re-stratified data improved fairness in some cases, like BERT (gender) and RoBERTa  (religion), but was not as effective as removing overamplification bias. We speculate that this is the case because removing selection bias includes re-stratifying the data by adding synthesized positive examples to the training dataset, leading to a balanced class ratio between positive and negative examples ($\approx 0.5$) for all identity groups. This resulted in the model predicting more false positives and less true negatives, which resulted in worse fairness and worse performance. On the other hand, training the model on perturbed data ensured balanced positive class representation between the different identity groups, but the ratio between the positive and negative class stayed low ($\approx 0.1 \text{ to } 0.2$). This made the model predict more positives, specifically more true positives, without hurting the number of true negatives, which improved both the performance and fairness of the inspected language models.

Removing both selection and overamplification bias (Downstream-perturbed-stratified-data), was more effective than removing only selection bias and less effective than removing only overamplification bias. Removing all sources of bias (upstream and downstream) resulted in the same pattern as fine-tuning the model on re-stratified perturbed data (+ downstream-perturbed-stratified-data), which confirms that removing upstream bias does not have a strong impact on the models' fairness. However, results showed that in rare cases, removing both upstream and downstream bias leads to improved fairness. For example, the FPR\_gap score for the gender-sensitive attribute of the AlBERT model (+downstream-perturbed-stratified-data) is 0.005, while the FPR\_gap score for the gender-sensitive attribute of the AlBERT model (+ upstream-sentDebias-gender-downstream-all-data-debias) is 0.001. However, this was not always the case.


\subsection{Which debiasing techniques to use to ensure the models' fairness on the task of toxicity detection?}
\label{sec:counterfactual_fairness}
The answer to the previous research questions suggests that the best technique to ensure the fairness of the task of toxicity detection is to remove overamplification bias by fine-tuning the models on a perturbed dataset. To confirm that this is indeed the most effective downstream debiasing technique and to answer this research question, we used the counterfactual fairness metric \cite{kusner2017counterfactual} to get a more profound understanding of the effect of the different bias removal techniques on the models' fairness. To this end, we used the perturbation sensitivity score ($SenseScore$) metric proposed by \citet{prabhakaran-etal-2019-perturbation} to inspect closely the models' prediction probabilities of the different identity groups within each sensitive attribute. $SenseScore$ is defined in Equation~\ref{eq:sens-scores} as the absolute mean of the difference between the model prediction ($f(x)$) of the factual ($x$) and the counterfactual (perturbed) ($\hat{x}$) instances, divided by the number of examples ($X$). For a model $f(x): X \rightarrow Y$, a factual instance ($x$) that contains identity ($g$), and counterfactual instance ($\hat{x}$) that contain identity ($\hat{g}$), fairness is measured as:
\begin{equation}
\label{eq:sens-scores}
    SenseScore = \left |Mean_{x \in X}(f(\hat{x}) - f(x)) \right |
\end{equation}

\begin{table}
    \renewcommand{\arraystretch}{1.30}
    \resizebox{\textwidth}{!}{
\begin{tabular}{l|l}
\hline
Sentence  & True label  \\ \hline
\begin{tabular}[c]{@{}l@{}}\% of \textcolor{orange} {men} lie to law-enforcement officers when question into domestic violence cases it is not without marks on the body or \\ conclusive proof that the \textcolor{orange} {men} are charged you \textcolor{orange} {sir} you are one of the \textcolor{orange} {men} that is the problem I can tell !\end{tabular}                  & 0   \\ \hline
\begin{tabular}[c]{@{}l@{}}\% of \textcolor{magenta} {women} lie to law-enforcement officers when question into domestic violence cases it is not without marks on the body or\\  conclusive proof that the \textcolor{magenta} {women} are charged you \textcolor{magenta} {mademoiselle} you are one of the \textcolor{magenta} {women} that is the problem I can tell !\end{tabular} & 0  \\ \hline
\end{tabular}}
\caption{Example of a sentence where the original target is a Male (top) and when the gender is swapped to Female (bottom).}
\label{tab:gendr-swap-example}
\end{table}
$SenseScore$ is an indicator of how the model treats different groups of people, since the sentence is the same with only the identity group being different. The bigger the score, the less fair the model is, since it means that the model treats the different groups differently. On the contrary, the smaller the score, the more fair the model is, since it means that it does not discriminate between the different groups of people based on sensitive attributes. This analysis is possible because the balanced civil community fairness dataset (section~\ref{sec:balanced_toxiciy_fairness_dataset}) contains counterfactual/perturbed examples. Consequently, when we measured the $SenseScore$ of two identity groups, e.g., Male and Female, we are actually measuring the difference of the models' prediction probabilities between the same sentences with only the gender identity keywords being different. This analysis was conducted for the different downstream debiasing techniques that proved the most effective and improved the models' fairness without hurting the models' performance, which are: re-stratification (Downstream-stratified-data), perturbation (Downstream-perturbed-data), and re-stratification and perturbation (Downstream-perturbed-stratified-data).

\begin{table}
\centering
    \renewcommand{\arraystretch}{1.2}
    \resizebox{0.8\textwidth}{!}{
    \begin{tabular}{l|r|r|r}
    \hline
         & \multicolumn{3}{c}{SenseScore}  \\ \hline
        Model & Gender & Race & Religion \\ \hline
        \textbf{AlBERT-base}                                  & $6.9 e^{-05}$ & 0.032 & 0.006 \\ \hline
        + downstream-perturbed-data             & \color{teal}$\downarrow4.2 e^{-05}$ & \color{teal}$\downarrow0.002$ & \color{teal}$\downarrow0.001$ \\ \hline
        + downstream-stratified-data            & \color{red}$\uparrow0.042$ & 0.032 & \color{red}$\uparrow0.009$ \\ \hline
        + downstream-perturbed-stratified-data & \color{red}$\uparrow0.013$ & \color{teal}$\downarrow0.003$ & \color{teal}$\downarrow0.0007$ \\ \hline

        \textbf{BERT-base} & 0.001 & 0.03 & 0.001 \\ \hline
        + downstream-perturbed-data & \color{teal} $\downarrow0.0007$ & \color{teal} $\downarrow0.003$ & 0.001 \\ \hline
        + downstream-stratified-data & \color{red} $\uparrow 0.025$ & \color{teal} $\downarrow0.022$ & \color{red}$\uparrow0.004$ \\ \hline
        + downstream-perturbed-stratified-data & \color{red}$\uparrow0.002$ & \color{teal} $\downarrow0.002$ & \color{teal} $\downarrow0.0008$ \\ \hline

        \textbf{RoBERTa-base} & 0.001 & 0.024 & 0.003 \\ \hline
        + downstream-perturbed-data & \color{teal}$\downarrow0.0008$ & \color{teal} $\downarrow0.006$ & \color{teal} $\downarrow0.001$ \\ \hline
        + downstream-stratified-data &  \color{red} $\uparrow 0.038$ & \color{red} $\uparrow 0.036$ & 0.003 \\ \hline
        + downstream-perturbed-stratified-data &  \color{red} $\uparrow 0.003$ & \color{teal}$\downarrow0.002$ & \color{teal}$\downarrow 0.0003$ \\ \hline
        \end{tabular}}
    \caption{SenseScores of the difference models before and after the different debiasing methods. (\textcolor{red}{$\uparrow$}) denotes that the fairness metric score increased and the fairness worsened. (\textcolor{teal}{$\downarrow$}) denotes that the fairness metric score decreased and the fairness improved.}
    \label{tab:sensescores}
\end{table}

We then examined the difference in $SenseScores$ of the examined debiasing techniques for the different sensitive attributes. For the gender attribute, we studied the sentences that are targeted at the Male group and that are perturbed to change the identity to the Female group, as shown in \tablename~\ref{tab:gendr-swap-example}, as well as the sentences that are targeted at the Female group and are perturbed to change the identity to the Male group. Then, we measured the $SenseScore$ between the same sentences with the Male and the Female identities swapped. For the race attribute, we examined the sentences that are targeted at the Black group and that are perturbed to change the identity to the White group, as well as the sentences that are targeted at the White group and are perturbed to change the identity to the Black group. For the religion attribute, we examined the sentences that are targeted at the Christian group and are perturbed by changing the identity to the Muslim group, as well as the sentences that are targeted at the Muslim group and are perturbed to change the identity to the Christian group.

The prediction sensitivity scores ($SenseScore$) shown in \tablename~\ref{tab:sensescores} indicate that removing overamplification bias is the most effective debiasing method. Fine-tuning the different models on a perturbed balanced dataset (+ downstream-perturbed-data) improved the fairness (lower $SenseScore$) for almost all the sensitive attributes, as evidenced by the results for AlBERT (gender, race, religion), BERT (gender, race), and RoBERTa  (gender, race, religion). The next most effective debiasing method is removing both selection and overamplification bias, since fine-tuning the different models on a perturbed-re-stratified balanced dataset (+ downstream-perturbed-stratified-data) improved the fairness for all the models, but only for the race and the religion sensitive attributes, as evidenced by the results for AlBERT (race, religion), BERT (race, religion), and RoBERTa  (race, religion). On the other hand, removing only the selection bias by fine-tuning the models on re-stratified data (+ downstream-stratified-data) is the least effective on the models' fairness, as it improved only the fairness of BERT (race) and deteriorated the fairness of AlBERT (gender, religion), BERT (gender, religion), and RoBERTa  (gender, race).


\section{How to improve fairness of Toxicity detection}
\label{sec:improving_fainress}
Now we draw from our results and findings in this paper to answer our last research question and to provide a list of guidelines to ensure the fairness of the task of toxicity detection.

\subsection{Guidelines}
\label{sec:guidlines}
Based on the findings of this work, we recommend a list of steps to follow to ensure the fairness of the downstream task of toxicity detection, as examined in this work in the context of toxicity detection. 

\begin{enumerate}
\setlength\itemsep{1em}
\item \textbf{Know the data:} The first recommendation is to know your data. This recommendation is the first step in any NLP task. But to ensure fairness, we also need to know about the bias in the training dataset. Especially since the results in \tablename~\ref{tab:source_bias_vs_fairness} indicate that downstream sources of bias are the most influential on the models' fairness. We recommend measuring the selection bias and overamplification bias in the training dataset.

In addition to measuring selection and overamplification bias, it is important to generally, investigate annotators bias if available and to check the data collection process and other data properties as argues by \citet{10.1145/3531146.3533231}. We also recommend learning about the historical and social context of the collected data to avoid undesirable societal impact \cite{benjamin2019race}.

\item \textbf{Remove overamplification bias:}  We recommend starting with removing the overamplification bias since it is the most impactful debiasing method on the models' fairness, as explained in section~\ref{sec:counterfactual_fairness}. We recommend removing overamplification bias by fine-tuning the model on the perturbed dataset to ensure the balanced contextual representations of the different identity groups, as shown in section~\ref{sec:extrinsic_bias}.  In addition to fine-tuning the model on balanced datasets, we recommend investigating different bias removal techniques to make sure that the right method is used with the right dataset.

\item \textbf{Know the model:} Similar to the data, it is important to know about the bias in the models being considered for the downstream task. Especially since the results suggest that there is a positive correlation between representation bias and the models' fairness. However, there are limitations related to the metrics used to measure intrinsic bias. We recommend using more than one metric. It is also important to check the intended use of the language model by checking the models' cards \cite {10.1145/3287560.3287596}, if available. 

\item \textbf{Balance the fairness data:} We recommend creating a perturbed version of the fairness dataset to make sure that the fairness dataset does not contain the selection or overamplification bias and that the measured fairness is more reliable, as explained in section~\ref{sec:extrinsic_bias}. 
  
\item \textbf{Measure counterfactual fairness:} Since the different fairness metrics provide different results and sometimes are not in agreement, we recommend using counterfactual fairness metrics. Especially since after balancing the fairness dataset, we have perturbed data items. This allows us to reliably measure how the model discriminates or not between the different groups of people. 

\item \textbf{Select the final model:} After collecting information about the bias in the fine-tuning dataset and the used language model and applying different debiasing methods to remove overamplification bias and measure fairness using counterfactual fairness metrics, we recommend to use the language model that achieves a good trade-off between performance and fairness.
\end{enumerate}
To further assess the generalizability of these guidelines for the general task of supervised text classification, we applied these guidelines to ensure the fairness of the task of sentiment analysis, as discussed in Appendix~\ref{sec:appendix}, showing that these guidelines are generalizable. However, further analysis and experiments on larger datasets are needed, and we leave that for future work.

\section{Limitations}
It is important to point out that this work is limited to the examined models and datasets. Bias and fairness are studied from a Western perspective regarding language (English) and culture. In addition, some concerns have been raised regarding how the different bias and fairness metrics articulate what the different metrics actually measure, as well as regarding the quality of the crowd-sourced datasets used to measure the bias \cite{Blodgett-etal-2021-norweigan-salmon}.
Besides, those metrics measure the existence of bias, not its absence, so a lower score does not necessarily mean that the model is unbiased \cite{DBLP:conf/naacl/MayWBBR19}. Nevertheless, when used for comparing different models, they can indicate whether a model is less biased than another.
Additionally, we were not able to replicate the representation bias scores reported in \citet{nangia-etal-2020-crows} and \cite{nadeem-etal-2021-stereoset}, because the latest version of the Transformer's Python package that we used (version 4) led to different results compared to the one used by the authors (version 3). The same finding was made by \citet{schick2021self}. 

Despite the fairness metrics used in this work being the most commonly used ones in the literature, they are not without criticism \cite{hedden2021statistical}. For example, \citet{hedden2021statistical} argues that group fairness metrics are based on criteria that cannot be satisfied unless the models make perfect predictions or that the base rates \cite{BARHILLEL1980211} are equal across all the identity groups in the dataset. 


Furthermore, we recognize that the provided recommendations to have a fairer toxicity detection task rely on creating perturbations for the training and the fairness dataset, and we acknowledge that this task may be challenging for some datasets, especially if the mention of the different identities is not explicit, like, for example, not using the word \say{Asian} to refer to an Asian person but using Asian names instead. 

Moreover, in this work, we aim to achieve equity in the fairness of the task of toxicity detection between the different identity groups. However, equity does not necessarily mean equality, as explained in \citet{Broussard2023}.

\section{Conclusion}
This work examined the impact of three different sources of bias, Representation, Selection and Overamplification bias, on the fairness of the downstream task of toxicity detection. This study covers three sensitive attributes: gender, race, and religion, as well as three widely used language models (BERT-base, ALBERT-base, and RoBERTa-base). 
We used metrics from the literature to measure representation bias. And we proposed methods to measure selection and overamplification bias. We, then, measure the impact of the different sources of bias on the fairness of toxicity detection, and evaluated the impact of bias removal on improving the models' fairness and performance.
Results showed that using a fairness dataset with the same contextual representation and ratio of positive examples for the different identity groups is a crucial step in measuring fairness. Unlike the findings of earlier research that suggested that there is no correlation between representation bias and the models' fairness, results show a consistent positive correlation between the representation bias scores measured by the CrowS-Pairs metric and fairness scores measured using different fairness metrics and when a balanced fairness dataset is used to measure fairness.

Additionally, our results consistently confirm that downstream sources of bias (selection and overamplification) are more impactful than upstream sources (representation bias), which is in line with other researchers' findings. Overamplification bias was shown to be the most impactful source of bias on the models' fairness, and removing it improved the fairness of the different models on the task of toxicity detection. Fine-tuning the models on the perturbed dataset led to the models providing similar prediction probabilities to the sentences where only the identity group is swapped, which suggests that the models do not discriminate between the different groups of people within the same identity group. 

Finally, the findings of this work were used to devise a set of guidelines for ensuring the fairness of the downstream task of toxicity detection, in the hopes that they would assist researchers and data scientists in creating fairer toxicity detection models.



\appendix

\appendixsection{Sentiment analysis}
\label{sec:appendix}

 
 To train a sentiment analysis model that is fair, we first need a sentiment analysis dataset that contains information about sensitive attributes. Since, to the best of our knowledge, there is no sentiment analysis dataset that contains identity information, we filtered the IMDB sentiment dataset\footnote{\url{https://www.kaggle.com/datasets/lakshmi25npathi/imdb-dataset-of-50k-movie-reviews}} to ensure that it contains gender information similar to the work in \citet{Webster-etal-2020-cda}. The keywords from \tablename~\ref{tab:gendered_keywords} were used to filter the IMDB dataset to ensure that gendered information is present in the training dataset. A gender column was then added to the dataset, and sentences were labelled as \say{Female} or \say{Male} based on the gendered keywords. There are cases when none of the identity words were found or a mixture of the words was found in the same sentence. In such cases, these sentences were labelled as \say{Neutral}. The IMDB dataset, after the keyword filtering, contained 50K data items. However, we then selected only the data items labelled as \say{Male} or \say{Female}. We call the filtered dataset \say{IMDB-gendered} and it contains 9,790 data items. 72\% of the data is labelled as ``Male'' and 27\% is labelled as ``Female''. The ratio of the positive examples in the subset of sentences that contain the Male identity is 0.55 and the ratio of positive examples in the subset of sentences that contain the Female identity is 0.52. IMDB-gendered was then randomly split as 40\% for training, 30\% for validation, and 30\% for the test, preserving the class ratio. Finally, three models were trained on the IMDB-gendered dataset, AlBERT-base, BERT-base, and RoBERTa-base, achieving AUC scores of 0.899, 0.912, and 0.914, respectively.

 \begin{table}[t]
     \centering
     \renewcommand{\arraystretch}{1}
     \resizebox{1\textwidth}{!}{
 \begin{tabular}{l|l}
 \hline
 Female  & \begin{tabular}[c]{@{}l@{}}she, her, hers, mum, mom, mother, daughter, sister, niece, aunt, grandmother,\\              lady, woman, girl, ma'am, female, wife, ms, miss, mrs, ms., mrs.\end{tabular} \\ \hline
 Male    & \begin{tabular}[c]{@{}l@{}}he, him, his, dad, father, son, brother, nephew, uncle, grandfather, gentleman, \\ man, boy, 'ir, male, husband, mr, mr.\end{tabular}                                      \\ \hline
 Neutral & they, them, theirs, parent, child, sibling, person, spouse                                                                                                                                            \\ \hline
 \end{tabular}}
 \caption{Keywords used to filter out the gendered sentences in IMDB dataset.}
 \label{tab:gendered_keywords}
 \end{table}



To measure gender fairness in the downstream task of sentiment analysis, we need a fairness dataset where the target of the sentiment in the sentence is identified. For this task, the SST-sentiment-fairness dataset \cite{sst-sentiment-fainress-dataset} was used. The dataset is a subset of the SST dataset that contains 462 data items with the target of the sentiment labelled by 3 annotators with an inter-annotation agreement of 0.65 and a Fleiss's Kappa score of 0.53. 42\% of the dataset is targeted at women (ratio of positive examples = 0.61) and 58\% is targeted at men (ratio of positive examples= 0.5).
The three sentiment analysis models that we fine-tuned on the IMDB-gendered dataset were used to predict the labels of the SST-sentiment-fairness dataset and to  measure their fairness, achieving AUC scores of 0.865 for AlBERT-base, 0.860 for BERT-base, and 0.878 for RoBERTA-base. 

We then applied the fairness guidelines described in section~\ref{sec:guidlines} to the sentiment analysis task as follows:
\begin{enumerate}
\setlength\itemsep{1em}
    \item \textbf{Know the data:} When we applied this to the IMDB-gendered dataset, using the methods proposed in sections \ref{sec:selection_bias} and \ref{sec:overap_bias}, we found that the selection bias score wass 0.03 and overamplification bias score was 0.309\footnote{Overamplification bias here is measured as the difference between the size of the male and female subsets divided by the size of the dataset.}

    \item \textbf{Remove overamplification bias:} We applied this technique to the IMDB-gendered dataset, and then, fine-tuned the models, AlBERT, BERT, and RoBERTa, on the perturbed-IMDB-gendered dataset. The AUC scores of the models on the perturbed-IMDB-gendered dataset are 0.869, 0.860, and 0.877 for AlBERT, BERT, and RoBERTa respectively.

    \item \textbf{Know the model:} Since we already measured the representation bias in section~\ref{sec:intrinsic_bias}, we do not need to repeat that for the sentiment analysis case study. 

    \item \textbf{Balance the fairness data:} We used the same method used in section~\ref{sec:extrinsic_bias} to create the gendered perturbed SST-fairness dataset. 50\% of the dataset is targeted at women (ratio of positive examples = 0.54) and 50\% is targeted at men (ratio of positive examples = 0.54). The data after perturbation contained 924 data items, with selection bias and overamplification bias both being 0.

    \item \textbf{Measure counterfactual fairness:}For the sentiment analysis task, we measured the fairness of the models after being fine-tuned on the perturbed IMDB-gendered dataset. We measured the counterfactual fairness of the models AlBERT, BERT, and RoBERTa on the perturbed SST-sentiment-fairness dataset, as explained in section~\ref{sec:counterfactual_fairness}. We found that the counterfactual fairness, as measured using the $Sensescore$ of the different models after removing Overamplification bias, is 0.005 for AlBERT, 0.0005 for BERT-base, and 0.009 for RoBERTa-base. 

    \item \textbf{Select the final model:} The results indicate that after removing overamplification bias, the model that discriminates the least between the male and the female groups in our sentiment analysis task is BERT (couterfactual fairness of 0.0005). Consequently, we chose BERT as it is the fairest model to use. This decision comes with a trade-off with regards to the performance score.
    On the perturbed-gendered-IMDB dataset, RoBERTa (+ downstream-perturbed-data) achieves an AUC score of 0.877, which slightly outperforms BERT (+ downstream-perturbed-data) that achieved an AUC score of 0.860. 
\end{enumerate}

\begin{acknowledgments}
This research has been partially supported by project ``AGENCY'', funded by the Engineering and Physical Sciences Research Council [EP/W032481/1], United Kingdom. Part of this work was done when the first author did an internship at IBM research in New York, the US.
\end{acknowledgments}

\bibliographystyle{compling}
\bibliography{bib.bib}

\end{document}